\documentclass[runningheads]{llncs}

\usepackage{eccv}

\usepackage{eccvabbrv}

\usepackage{graphicx}
\usepackage{booktabs}

\usepackage[accsupp]{axessibility}  

\usepackage{multibib}
\newcites{supp}{Supplementary References}

\usepackage{hyperref}

\usepackage{orcidlink}

\usepackage{cancel}

\usepackage{cite}
\usepackage{graphicx}
\usepackage[font=small,labelfont=bf]{caption}
\usepackage[font=small,labelfont=bf]{subcaption}
\usepackage{float}
\usepackage[T1]{fontenc}
\usepackage{multirow}
\usepackage{enumerate}
\usepackage{amsmath}
\usepackage{adjustbox}
\usepackage{threeparttable}
\usepackage{wrapfig}
\usepackage{xspace}
\usepackage{xcolor}   
\usepackage{pifont}    

\definecolor{formalgreen}{rgb}{0.1, 0.7, 0.1}  
\definecolor{formalred}{rgb}{0.9, 0.2, 0.2}  
\newcommand{\cmark}{\textcolor{formalgreen}{\checkmark}}  
\newcommand{\xmark}{\textcolor{formalred}{\ding{55}}}           
\usepackage[dvipsnames]{xcolor}

\definecolor{lightblue}{RGB}{100, 149, 237}
\definecolor{green2}{rgb}{0, 0.7, 0.3}
\definecolor{orange2}{rgb}{1.0, 0.8, 0.2}

\newcommand{\red}[1]{{\color{red}#1}}
\newcommand{\green}[1]{{\color{green2}#1}}
\newcommand{\orange}[1]{{\color{orange2}#1}}
\newcommand{\blue}[1]{{\color{blue}#1}}
\newcommand{\lightblue}[1]{{\color{lightblue}#1}}

\usepackage{colortbl}
\usepackage{multirow}
\usepackage[para]{footmisc}
\usepackage{lipsum}
\usepackage{booktabs}

\usepackage{xspace}
\usepackage{hyperref}

\makeatletter
\DeclareRobustCommand\onedot{\futurelet\@let@token\@onedot}
\def\@onedot{\ifx\@let@token.\else.\null\fi\xspace}

\def\eg{\emph{e.g}\onedot}

\makeatother

\usepackage{amsmath}
\usepackage{amssymb}
\usepackage{dsfont}
\usepackage{etoolbox}
\usepackage{color}
\usepackage{ulem}

\newif\ifshowedits

\newcommand{\addeditor}[3]{%
  \definecolor{#1color}{rgb}{#3}
  \expandafter\newcommand\csname #1\endcsname[1]{%
  \ifshowedits
    {\color{#1color} ##1}%
  \else
    {##1}%
  \fi
  }%
  \expandafter\newcommand\csname #1rmk\endcsname[1]{%
  \ifshowedits
    {\color{#1color} {\bf [#2: ##1]}}
  \fi
  }%
  \expandafter\newcommand\csname #1rpl\endcsname[2]{%
  \ifshowedits
    {\color{#1color} ##1 \sout{##2}}
  \else
    {##1}
  \fi
  }%
}

\newcommand{\createtextvar}[1]{
  \expandafter\newcommand\csname #1\endcsname{%
  {\text{#1}}
}%
}
\newcommand{\mycomment}[1]{}

\newcommand{\calA}{{\cal A}}

\newcommand{\calC}{{\cal C}}
\newcommand{\calD}{{\cal D}}

\newcommand{\calF}{{\cal F}}

\newcommand{\calH}{{\cal H}}
\newcommand{\calI}{{\cal I}}
\newcommand{\calJ}{{\cal J}}

\newcommand{\calN}{{\cal N}}
\newcommand{\calO}{{\cal O}}
\newcommand{\calP}{{\cal P}}

\newcommand{\calS}{{\cal S}}
\newcommand{\calT}{{\cal T}}
\newcommand{\calU}{{\cal U}}

\newcommand{\calW}{{\cal W}}

\newcommand{\calZ}{{\cal Z}}

\newcommand{\vcomment}[1]{}

\addeditor{julien}{JM}{1.0, 0.349, 0.0}
\addeditor{boris}{BM}{0.0, 0.5, 0.0}
\addeditor{mathieu}{MG}{0,0,0.5}
\addeditor{liming}{LC}{0.7,0.2,0.8}
\addeditor{template}{TP}{1,0,0}
\showeditstrue

\newcommand{\acronyme}{CoToGrasp\xspace} 
\newcommand{\acronymefull}{Contact-Topology-Conditioned Dexterous Grasp Synthesis via Canonical Workspace Learning\xspace}

\begin{document}

\title{\acronyme: \acronymefull} 

\titlerunning{CoToGrasp}

\author{Julien Mérand\inst{1}\orcidlink{0009-0005-0012-2985} \and
Boris Meden\inst{1}\orcidlink{0000-0003-4521-2595} \and
Liming Chen\inst{2}\orcidlink{0000-0002-3654-9498} \and
Mathieu Grossard\inst{1}}

\authorrunning{J.~Mérand et al.}

\institute{Université Paris-Saclay, CEA, List, F-91120 Palaiseau, France \and
Ecole Centrale Lyon, CNRS, LIRIS, UMR5205, Institut Universitaire de France (IUF), F-69130 Ecully, France}

\maketitle

\begin{abstract}

Current dexterous grasp planners primarily optimize for physical stability, focusing on whether an object can be grasped rather than how it should be grasped to support downstream functional tasks. However, conditioning grasp synthesis on specific human grasp taxonomies typically requires prohibitively expensive, object-annotated datasets.
To address these limitations, we propose \acronyme, a novel generative framework that synthesizes diverse, stable grasps strictly conditioned on specific contact topologies. To bypass the data collection bottleneck, \acronyme is trained entirely in an object-agnostic manner. We introduce a feature-based canonical workspace that projects local object features into a unified gripper-centric domain, effectively decoupling the semantic functional intent from the arbitrary object geometry. By learning the intrinsic contact manifold of the gripper within this workspace, our model achieves zero-shot generalization to unseen objects at inference. Extensive evaluations on the large-scale DexGraspNet dataset demonstrate that \acronyme achieves state-of-the-art performance, outperforming existing taxonomy-guided planners. Finally, we demonstrate the physical viability and kinematic feasibility of our synthesized contact topologies on a physical robot platform. Code is available on our project website \blue{\url{https://cea-list.github.io/cotograspweb/}}.

\end{abstract}

\section{Introduction}

\begin{figure}[th]
    \centering
    \includegraphics[width=\columnwidth]{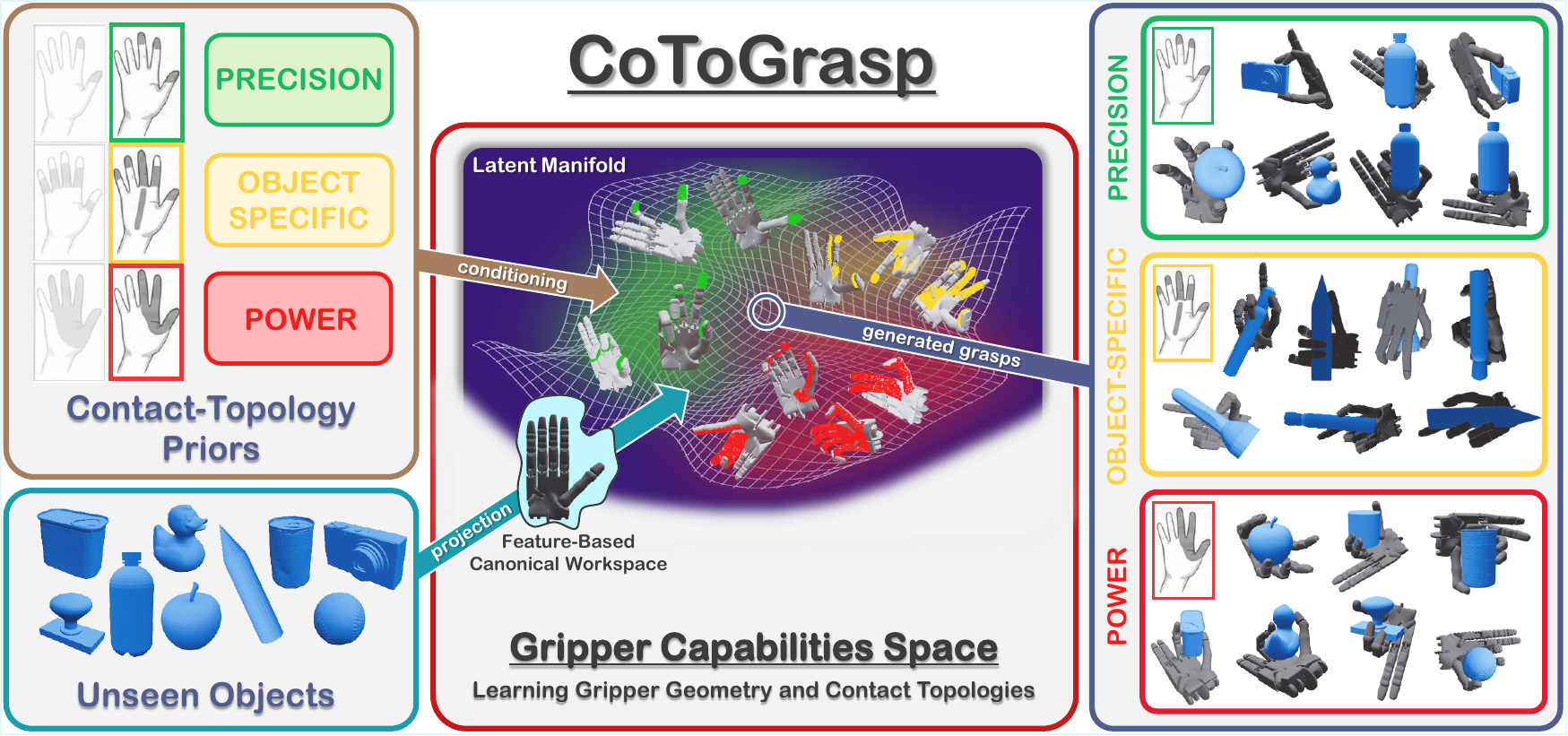}
    \caption{\textbf{Contact-Topology-Conditioned Grasp Synthesis.} Given a desired semantic contact-topology condition (top left) -- categorized into \textbf{\green{Precision}}, \textbf{\orange{Object-Specific}} (highly constrained topologies tailored for specific tool use) or \textbf{\red{Power}} functional groups -- and a novel, unseen object (bottom left), our framework synthesizes functionally diverse and physically stable grasps (right). Rather than learning contact topologies directly on the object geometry, we project local object features into a feature-based canonical workspace. This unified spatial representation effectively decouples the functional intent from the specific object identity. Within this workspace, we learn a latent manifold (center) that models the intrinsic contact capabilities of the gripper, enabling zero-shot generalization to diverse target geometries.}
    \label{fig:teaser}
\end{figure}

As robotic systems evolve toward embodied agents capable of complex interaction, grasping can no longer be treated solely as a geometric stability optimization problem. The recent rise of humanoid robots, increasingly guided by high-level reasoning frameworks such as Large Language Models (LLMs)~\cite{kim2024survey}, demands grasps that directly answer functional, task-level intents. Fine-grained robot manipulation necessitates dexterous, multi-fingered hands with high Degrees of Freedom (DoF) because many downstream tasks -- such as in-hand object reorientation, precision tool insertion, finger gaiting, and handle-based grasping -- demand controllable, distributed multi-point contact topologies that fundamentally exceed the symmetric pinch and enveloping capabilities of simple parallel-jaw grippers.

However, the transition to high-DoF systems exposes a critical limitation in existing grasp planners: while they can produce physically stable grasps, they lack the structural properties required to generate task-aligned contact topologies. Modern data-driven grasp planners predominantly focus on whether an object can be grasped rather than how it should be manipulated. By optimizing purely for geometric stability and force closure, these methods introduce a severe generative bias toward energetically stable but functionally uniform grasps. Although dexterous hands offer rich articulated contact capabilities, current planners do not fully exploit their potential in terms of grasp pattern diversity, instead overwhelmingly defaulting to enveloping power grasps. Consequently, synthesizing a specific precision grasp required for a downstream task becomes highly inefficient, as it necessitates the generation and rejection of an impractical volume of functionally unsuitable candidates.

To achieve task-aligned grasp synthesis, it is necessary to introduce a structural prior over valid contact configurations. In this work, rather than relying on arbitrary stable grasps, we condition grasp synthesis on structured contact topologies derived from human grasp taxonomies. Specifically, we adapt the Gonzalez taxonomy~\cite{gonzalez2014analysis}, which categorizes grasps based strictly on the hand's active contact surfaces rather than the object's shape. This hand-centric formulation allows for seamless adaptation to anthropomorphic grippers independent of their specific kinematics. Furthermore, to avoid the severe generalization bottlenecks associated with explicitly mapping contact topologies to specific object geometries in training datasets, we propose building on the object-agnostic training paradigm introduced in \cite{merand2026goag}. By learning the intrinsic contact capabilities of the robotic hand independently of specific objects, we completely decouple grasp semantics from object topology.

We introduce \acronyme (\acronymefull), a gripper-oriented framework for generating diverse, contact-topology-conditioned grasps. Unlike standard methods that learn contact maps directly on object point cloud, our approach utilizes a Canonical Feature-based Workspace anchored to the gripper frame. 
This workspace acts as a domain-agnostic bridge: we first extract local geometric features via a DGCNN~\cite{phan2018dgcnn} encoder from either the gripper (training) or the object (inference), then aggregate these features into fixed points within the canonical workspace using k-Nearest Neighbors (kNN).
To capture the complex spatial structure of functional grasps, we treat each populated workspace point as a discrete token and process this set of workspace-anchored embeddings using a Transformer~\cite{vaswani2017attention} encoder. This attention mechanism allows the network to learn structural constraints by modeling the specific spatial distribution and co-occurrence patterns of contact points inherent to each topology.
These refined features are distilled into a compact latent descriptor via a Set Transformer~\cite{lee2019set}, which conditions a Conditional Variational Auto-Encoder (CVAE)~\cite{sohn2015learning} to reconstruct probabilistic contact masks. Finally, we synthesize the physical grasp through a test-time energy-based optimization that aligns the gripper’s active surface with the predicted contact zones, enforcing kinematic constraints and collision avoidance.

We evaluate our method extensively in both simulation and real-world settings. We first demonstrate the severe functional bias of current taxonomy-unaware planners by measuring the entropy of their retrieved contact topology distributions. We then compare \acronyme against state-of-the-art taxonomy-guided methods, demonstrating superior topology compliance and stability.

In summary, our main contribution is the introduction of \acronyme, a novel object-agnostic grasp planner that synthesizes grasps conditioned on structured contact topologies derived from human taxonomies. Furthermore, to properly assess these capabilities, we establish a rigorous evaluation methodology to quantify the functional diversity and semantic bias inherent in dexterous grasp planners.
\section{Related Work}

\noindent\textbf{Learning-Based Grasp Planners.}
The rise of humanoid robots demands robust, task-oriented multi-fingered grasp synthesis~\cite{gu2026humanoid, li2025developments}.
However, contemporary data-driven grasp planners -- whether directly predicting explicit joint configurations~\cite{liu2020deep, xu2024dexterous, jiang2021hand, weng2024dexdiffuser, huang2023diffusion, lu2024ugg, xu2024manifoundation, zhong2025dexgrasp, wei2024grasp} or learning intermediate grasp representations~\cite{wei2024d, attarian2023geometry, li2023gendexgrasp, khargonkar2025robotfingerprint} -- prioritize physical stability over functional intent.
Heavily reliant on synthetic databases~\cite{wang2023dexgraspnet, turpin2023fast, xu2023unidexgrasp} generated via analytical force-closure resolution~\cite{liu2021synthesizing, miller2004graspit}, these models frequently suffer from mode collapse. They default to enveloping power grasps, underutilizing hand dexterity. Furthermore, explicitly mapping grippers to specific objects biases these models, hindering shape generalization. To circumvent this, frameworks like GOAG~\cite{merand2026goag} utilize object-agnostic paradigms, demonstrating that learning contact topologies directly within the gripper's canonical space yields superior generalization to novel geometries.

\noindent\textbf{Human Taxonomies and Task-Oriented Semantics.}
To address how an object should be grasped, researchers draw from biomechanics. Cutkosky~\cite{cutkosky1989grasp} established early categorizations based on object properties and task requirements, while Bullock et al.~\cite{bullock2012hand} considered the motion dynamics. Feix et al.~\cite{feix2015grasp} later proposed the comprehensive GRASP taxonomy. Crucially for haptics, Gonzalez et al.~\cite{gonzalez2014analysis} synthesized these frameworks by categorizing 21 distinct contact topologies based strictly on the hand's active contact surfaces. \\
Concurrently, task-oriented planners have leveraged these taxonomies to inject high-level semantics into robotic actions. Early studies utilized CNNs to decode manipulation semantics~\cite{deng2021adaptive, lu2020multifingered, rao2018learning}, while recent approaches employ Vision-Language Models~\cite{li2024multi, li2024semgrasp} to guide grasp selection.
However, bridging the gap between high-level language semantics and low-level physical interactions remains challenging. Methods like FunGrasp~\cite{huang2025fungrasp} retarget human-object interactions to grippers, but remain heavily dependent on task-specific interaction priors.

\noindent\textbf{Taxonomy-Conditioned Grasp Synthesis.} 
To explicitly control functional intent, recent methods condition their generative pipelines on taxonomy types. These representations vary from holistic couplings of joints, contacts, and object geometry~\cite{kleer2024incorporation, lu2019modeling}, to manual grouping~\cite{wu2024cross}, to purely kinematic "eigen grasps"~\cite{fang2025anydexgrasp, mao2026demofungrasp}. Recent frameworks like Dexonomy~\cite{chen2025dexonomy} and OmniDexVLG~\cite{zhang2025omnidexvlg} represent types via explicit joint and contact combinations.
Despite these advancements, current planners share a critical limitation: the deeply ingrained coupling of taxonomy types with specific object geometries during dataset construction. Methods like Dexonomy~\cite{chen2025dexonomy} require massive, analytically computed datasets where grasp types are strictly annotated against specific object meshes. If a novel object's local geometry does not perfectly accommodate the rigid pre-computed joint template, the optimization fails or collapses into a functionally incorrect type.
\acronyme subverts this limitation by utilizing a purely hand-centered taxonomy~\cite{gonzalez2014analysis} based on semantic contact masks rather than rigid joint configurations. By training our generative model in a strictly object-agnostic manner, we entirely bypass the need for costly, object-annotated datasets.
\section{Method}

\begin{figure*}[t]
    \centering
    \includegraphics[width=\columnwidth]{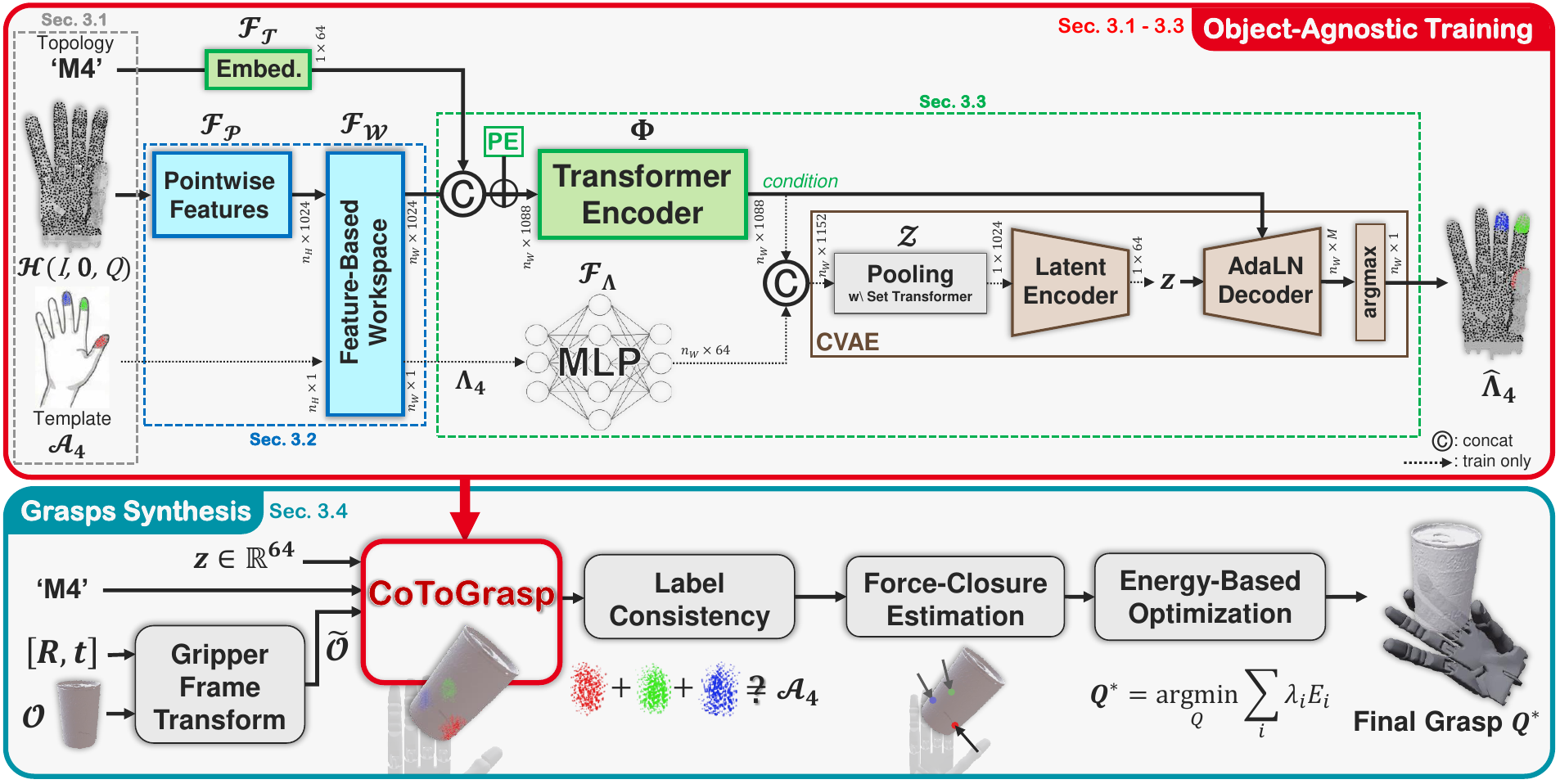}
    \caption{\textbf{\acronyme Method Overview.} 
    The proposed framework operates in two distinct phases. \textbf{Top (Object-Agnostic Training):} The model learns an intrinsic, gripper-centric contact manifold within a canonical feature-based workspace, independent of object geometry. \textbf{Bottom (Grasp Synthesis):} At inference, a target object is transformed into the canonical frame. The network's contact-topology-conditioned prediction is strictly filtered through a validation pipeline before energy-based optimization aligns the gripper to yield the final stable grasp ($Q^*$).
    }
    \label{fig:method_overview}
\end{figure*}

Figure~\ref{fig:method_overview} provides an overview of the \acronyme framework. 
Formally, we define a grasp as a tuple $(R, \mathbf{t}, Q)$, where $(R, \mathbf{t}) \in SO(3) \times \mathbb{R}^3$ represents the 6D pose of the gripper relative to the object frame $\calO$, and $Q$ denotes the internal joint configuration.
Our primary objective is to synthesize a diverse set of physically stable grasps that strictly respect a requested semantic contact topology.

To succesfully decouple high-level functional semantics from arbitrary object geometries, we adopt an object-agnostic learning paradigm. Unlike standard methods, our architecture is trained exclusively on gripper point clouds and inferred on target object point clouds.
During the training phase, the model operates entirely within the canonical gripper frame, rendering it independent of the global pose $(R, \mathbf{t})$. The network takes as input only the gripper's local surface geometry and a semantic contact topology, learning to reconstruct the corresponding physical contact template mask.

\subsection{Gripper-Oriented Contact Topology}
\label{subsec:taxonomy_formulation}

Let $\calI = \{1, \dots, N_H\}$ be the fixed index set of the points on the gripper's active grasping surface -- the specific subset of the gripper geometry designed to establish physical contact -- representing a configuration-independent domain. We define the gripper handprint, $\calH(R, \mathbf{t}, Q)$, as the spatial embedding of these points given a 6D pose $(R, \mathbf{t})$ and a joint configuration $Q$:
\begin{equation}
    \calH(R, \mathbf{t}, Q) = \{ \mathbf{h}_i(R, \mathbf{t}, Q) \in \mathbb{R}^6 \mid i \in \calI \}
\end{equation}
where $\mathbf{h}_i(R, \mathbf{t}, Q)$ denotes the global Cartesian coordinates $(x, y, z)$ of the $i$-th point concatenated with its corresponding surface normal $(n_x, n_y, n_z)$. The handprint $\calH$ is strictly defined as the discretization of the gripper's active grasping surfaces (detailed in Supp. Mat. A). 

\begin{figure}[t]
    \centering
    \includegraphics[width=\columnwidth]{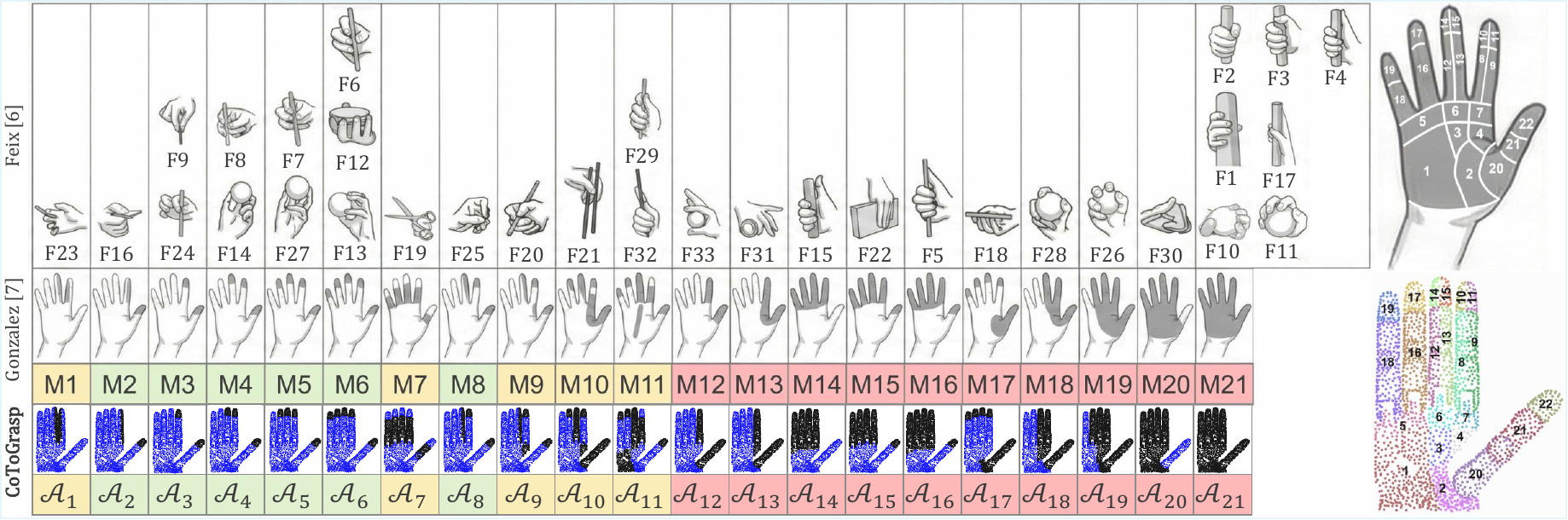}
    \caption{\textbf{Semantic Grasp Taxonomy and Contact Mapping.} \textbf{(Left)} Correspondences between the classical Feix~\cite{feix2015grasp} (F) taxonomy (top), the haptic Gonzalez~\cite{gonzalez2014analysis} (M) taxonomy (middle row) and our derived point cloud contact templates $\calA_m$ (bottom row). We categorized the 21 templates into three distinct functional groups: \textbf{\green{Precision}}, \textbf{\red{Power}} and \textbf{\orange{Object-Specific}} (highly constrained topologies tailored for specific tool use).
    \textbf{(Right)} The 22 anatomical contact zones defined by Gonzalez (top) and the direct surjective mapping ($\zeta$) onto our discrete gripper handprint $\calH$ (bottom).}
    \label{fig:grasp_taxonomy_contact}
\end{figure}

We adapt the taxonomy $\calT$ from~\cite{gonzalez2014analysis}, consisting of $M=21$ pairs of topology names and contact templates. Each template $\calA_m: \calI \longrightarrow \{0, \dots, N\}$ acts as a semantic mask, assigning a Zone ID to points required for contact topology $m$:
\begin{equation}
    \calT = \{(\text{Name}_m, \calA_m) \mid m\in[1,M] \}
    \text{, } \qquad
    \mathcal{A}_m(i) = \begin{cases} \zeta(i) & \text{if } i \in \calS_m \\ 0 & \text{otherwise} \end{cases}
\end{equation}
where $\zeta: \calI \to \{1, \dots, N\}$ is a surjective mapping associating every hand point, from $\calH$, to a physical zone and $\calS_m$ is the set of active points for contact topology $m$. This formulation, illustrated in Figure~\ref{fig:grasp_taxonomy_contact}, enables the systematic adaptation of the human-centric taxonomy~\cite{gonzalez2014analysis} to anthropomorphic grippers. Crucially, this taxonomy serves purely as a labeling interface. CoToGrasp easily adapts to any alternative contact-based representation.

\subsection{Geometric Transfer via Canonical Workspace Learning}
\label{subsec:canonical_workspace_learning}

\noindent\textbf{The Duality of Contact.}
Contact between a gripper and an object is fundamentally a symmetric geometric relation: a point on the gripper surface is in contact with the object if and only if a corresponding object point is in contact with the gripper.
\cite{li2023gendexgrasp, khargonkar2025robotfingerprint} formulate contact detection as an object-centric problem, parameterizing the search for valid grasp locations over the object surface $\calO$ to define the contact set $\calC_{\text{obj}} \subset \calO$. Conceptually, they ask, "where on the object can I touch?".
However, this duality allows us to invert the paradigm. Instead, we ask, "which regions of my hand are activated by this object?" By flipping the formulation, we define the gripper contact set $\calC_{\text{grip}}$ directly on the hand's surface. We reformulate the contact condition as:
\begin{equation}
    \calC_{\text{grip}}(\calH(R, \mathbf{t}, Q)) = \{ h_i \in \calH(R, \mathbf{t}, Q) \mid \min_{o \in \calO} \calD_{\text{aligned}}(h_i, o) < \epsilon \}
\label{eq:gripper_contact_def}
\end{equation}
with the distance $\calD_{aligned}$ defined between any two points $\mathbf{x}$ and $\mathbf{y}$, introduced in~\cite{li2023gendexgrasp}, as:
\begin{equation}
\label{eq:aligned_dist}
    \calD_{aligned}(\mathbf{x}, \mathbf{y}) = e^{\gamma(1-\langle \mathbf{x} - \mathbf{y}, n_{\mathbf{x}}\rangle)} \sqrt{\Vert \mathbf{x} - \mathbf{y} \Vert_2}
\end{equation}
where $n_{\mathbf{x}}$ the surface normal at $\mathbf{x}$ and $\gamma$ is a scaling factor modulating the influence of the kinematic alignment compared to the pure Euclidean distance.

This duality matters profoundly because it fundamentally changes the learning domain. The object surface $\calO$ represents an unbounded domain with infinite variability and high geometric entropy. In contrast, the gripper surface $\calH$ possesses a fixed structure, a bounded domain, and a finite kinematic manifold. 
By shifting the formulation from an unbounded object space to the fixed, mechanism-intrinsic gripper manifold, the solution space $\calC_{\text{grip}}$ is strictly bounded by $\calH$, simplifying the generative task. Consequently, the network avoids memorizing arbitrary target geometries, instead learning grasp templates purely as fundamental, object-agnostic capabilities of the gripper.

\noindent\textbf{Shift to a Gripper-Centric Frame.}
Conceptually, this duality allows us to invert the traditional learning paradigm by anchoring our perspective entirely on a canonical gripper frame, where the palm rests permanently at the origin ($R=I, \mathbf{t}=\mathbf{0}$). Here, the handprint's spatial embedding, $\calH(Q)$, depends strictly on the internal joint configuration $Q$, completely isolated from the global grasp pose. 
The target object $\calO$ is brought into this shared space via the inverse transformation $(R, \mathbf{t})^{-1}$, yielding $\tilde{\mathcal{O}}$. Crucially, because physical contact involves opposing surfaces (Eq.~\ref{eq:aligned_dist}), we invert the object's surface normals, transforming its local geometry into a negative mold of the expected gripper surface. 

\noindent\textbf{Feature-Based Workspace Representation.}
A central challenge arises from the fact that training and inference operate on geometrically distinct domains. Directly learning over these heterogeneous domains would entangle contact reasoning with object-specific topology. To resolve this, we introduce a canonical feature-based workspace $\calW = \{w_j \in \mathbb{R}^{3}\}_{j=1}^{N_W}$, defined as a fixed set of spatial basis points anchored to the gripper frame. \\
To capture fine-grained surface details, we extract a dense local description from the input geometry $\calP = \{\mathbf{p}_i \in \mathbb{R}^6\}_{i=1}^{N_P}$. We employ a modified DGCNN~\cite{phan2018dgcnn} to extract point-wise local features $\calF_{\calP} = \{\mathbf{f}_i\}_{i=1}^{N_P}$. \\
To disentangle these features from the input topology, we project them onto the fixed basis $\calW$ using a weighted $k$-Nearest Neighbors (kNN) aggregation. To ensure surface orientation strictly dictates contact viability, we gate the aggregation using the aligned distance ($\calD_{\text{aligned}}$). Let $\calN_j$ denote the indices of the $k$ nearest points in $\calP$ to the workspace basis point $w_j$. The aggregated feature $\calF_{\calW_j}$ is computed as:
\begin{equation}
\label{eq:features_projection}
    \calF_{\calW_j} = \frac{1}{k} \sum_{i \in \calN_j} \mathbf{f}_i \cdot e^{-\calD_{\text{aligned}}(\mathbf{p}_i, w_j)}
\end{equation}
By embedding local geometric features directly into the canonical workspace basis $\calW$, we construct a consistent set of spatial tokens, effectively decoupling the semantic contact reasoning from the arbitrary input geometry (further projection details and justifications are provided in Supp. Mat. B).

\noindent\textbf{Projected Ground Truth Templates.}
To construct the supervisory signal used during training, we project the templates $\calA_m$ onto the canonical workspace $\calW$. We strictly gate this projection using the distance threshold $\epsilon = 0.8$, consistent with our contact formulation in Eq.~\ref{eq:gripper_contact_def}. Each workspace point $w_j$ is assigned a label according to its nearest kinematically aligned hand point:
\begin{equation}
    \Lambda_m(j) = 
    \begin{cases}
    \calA_m(i^*) & \text{if } \calD_{\text{aligned}}(\mathbf{h}_{i^*}(Q), w_j) < \epsilon \\
    0 & \text{otherwise}
    \end{cases}
\end{equation}
where $i^* = \underset{i \in \calI}{\mathrm{argmin}} \ \calD_{\text{aligned}}(\mathbf{h}_i(Q), w_j)$. \\
Workspace points lying beyond this threshold are assigned the label $0$, indicating no physical contact. Consequently, $\Lambda_m \in \{0, \dots, N\}^{N_W}$ serves as the explicit ground truth contact map.

\subsection{Learning Contact Topology Templates using Self-Attention}
\label{subsec:self_attention}

To model non-local dependencies between potential contact regions, we process the workspace embeddings using a Transformer~\cite{vaswani2017attention} encoder. To prevent the semantic signal of the requested contact topology from diluting across deep attention layers, we explicitly concatenate the learnable topology embedding $\calF_{\calT}$ to every workspace point feature $\calF_{\calW}$, denoted $\calF_{\calW \oplus \calT}$. We inject spatial awareness via sinusoidal positional encodings (PE), allowing the multi-head attention mechanism to recover spatial relationships purely from the fixed basis indices:
\begin{equation}
    \Phi = \text{Transformer}(\calF_{\calW \oplus \calT} + \text{PE})
\end{equation}

To bridge the modality gap, the projected templates $\Lambda_m$ are passed through a Multi-Layer Perceptron (MLP) to create a continuous embedding $\calF_{\Lambda}$. These are fused with the transformer features $\Phi$ and compressed into a global latent descriptor $\calZ$ using a Set Transformer~\cite{lee2019set}. We leverage a Pooling by MultiHead Attention (PMA) block and a Set Attention Block (SAB) to capture complex structural dependencies:
\begin{equation}
    \calZ = \text{SAB}(\text{PMA}_s(\Phi \oplus \calF_{\Lambda}))
\end{equation}

Finally, a CVAE~\cite{sohn2015learning} models the localized, intra-topology stochasticity of the contact patches. Because macroscopic grasp multi-modality is explicitly resolved by the topology conditioning, a standard unimodal Gaussian prior $\calN(0,I)$ proves highly efficient. A convolutional encoder maps $\calZ$ to the posterior distribution $q(z|\calZ)$, from which a latent variable $z \in \mathbb{R}^{\psi}$ is sampled. The decoder, conditioned on $\Phi$ via Adaptive Normalization~\cite{xu2019understanding} layers (AdaLN), modulates the features to output the predicted zone probabilities $\hat{\Lambda}_m$.

The network is trained end-to-end to minimize the standard variational lower bound (ELBO), consisting of a Multi-Class Cross-Entropy reconstruction loss over the workspace points and a Kullback-Leibler regularization term weighted by $\beta$. (See Supp. Mat. B for architectural justifications and loss formulations.)

\subsection{Grasp Synthesis Through Contact Points Generation}
\label{subsec:inference}

In the inference phase, our goal is to synthesize a valid joint configuration $Q$ that realizes a specific contact topology $m$ for an object $\calO$ at a candidate pose $(R, \mathbf{t})$. We first transform the object point cloud into the canonical gripper frame via the inverse rigid transformation $(R, \mathbf{t})^{-1}$. This transformed geometry $\tilde{\calO}$, along with the requested contact topology $m$, is fed into the network -- bypassing the latent encoder by sampling $z$ from the prior $\calN(0,I)$ -- to predict a contact template $\hat{\Lambda}_m$ over the workspace. To ensure the physical viability of the grasp before costly optimization, we enforce a strict cascading validation pipeline:

\noindent\textbf{Label-Consistency Check.}
A contact topology is fundamentally constrained by the object's local geometry. To prevent generating ill-posed grasps, we apply a template-consistency filter immediately after inference. We perform a strict binary check on the symmetric difference between the ground-truth ($\Lambda_m$) and predicted ($\hat{\Lambda}_m$) templates. We allow a maximum deviation of one missing or hallucinated contact zone, safely rejecting fundamentally distinct failures where the object cannot support the requisite contacts.

\noindent\textbf{Contact Points Force-Closure Validation.}
For candidates passing the label-consistency check, we assess the potential grasp stability using the rapid, approximated force-closure estimation proposed in~\cite{merand2026goag}. Specifically, we extract the 3D workspace points $w_j \in \calW$ assigned a non-zero label by $\hat{\Lambda}_m$. We compute the barycenters of these active regions to construct the theoretical grasp wrench space. Crucially, if the force-closure condition is not met, it implies the proposed contact distribution is physically unviable. In this case, we discard the prediction and exploit the generative capacity of our network by resampling the latent variable $z$ to produce a fresh contact template $\hat{\Lambda}_m$. We limit this resampling process to a maximum of 20 iterations to prevent exhaustive searches in poorly conditioned poses. (Detailed formulations are provided in Supp. Mat. D).

\noindent\textbf{Joint Configuration Optimization.}
Finally, we retrieve the optimal joint configuration $Q^*$ by minimizing an energy function that aligns the gripper's active surfaces with the predicted 3D spatial workspace points, while respecting kinematic constraints:
\begin{equation}
\label{eq:joint_optimization}
    Q^* = \operatorname*{argmin}_{Q} \left( \lambda_d E_{dist} + \lambda_p E_{pen} + \lambda_s E_{spen} + \lambda_j E_{j} + \lambda_r E_{rep} \right)
\end{equation}
We adopt the standard terms $E_{dist}$ (contact alignment), $E_{pen}$ (object penetration), $E_{spen}$ (self-collision), and $E_{j}$ (joint limits) used in~\cite{merand2026goag, khargonkar2025robotfingerprint, li2023gendexgrasp}.
Additionally, to best match the desired contact topology without prior shape knowledge, we introduce a repulsive term $E_{rep}$. This term creates a repulsive field around the object for all unused fingers and palm regions (i.e., those not assigned a valid label in $\hat{\Lambda_m}$), enforcing a minimum safety margin of 5 mm to guide the optimizer toward strictly topology-compliant, collision-free configurations. (Detailed mathematical formulations and weighting factors are provided Supp. Mat. D).
\section{Experiments}
\label{sec:experiments}

\subsection{Experimental Setup}
\label{subsec:exp_setup}

\noindent\textbf{Training Data Formulation.} 
To construct the object-agnostic training dataset, we first define the semantic mapping between the physical surface regions of the target gripper and the required active contact zones ($N=22$) for the $M=21$ contact topology templates. We uniformly sample $10,000$ kinematically valid joint configurations $Q$ and compute their corresponding spatial handprints $\calH(Q)$ via forward kinematics. By pairing every sampled handprint with all 21 contact templates, we construct a comprehensive training dataset of 210,000 unique data points. Crucially, because this formulation operates entirely within the canonical gripper frame, dataset generation requires zero object meshes or computationally expensive physical simulations. (Further implementation and training details are provided in Supp. Mat. A).

\noindent\textbf{Grasp Pose Constraints and Generation.}
Because \acronyme operates in a canonical, gripper-centric workspace (Sec.~\ref{subsec:canonical_workspace_learning}), it inherently decouples the global 6D grasp pose $(R, \mathbf{t}) \in SO(3) \times \mathbb{R}^3$ from local contact generation. By treating this pose as an external prior rather than inferring it end-to-end, our architecture allows task planners or human operators to dictate functionally suitable approaches. This modularity facilitates advanced applications like kinematic keyframing for in-hand manipulation, enabling the synthesis of continuous, topology-consistent grasps along defined object trajectories.  \\
To autonomously evaluate \acronyme without external planners, we sample diverse candidate poses using a topology-conditioned heuristic depending on whether the requested topology involves the palm or relies strictly on distal precision. (Detailed sampling formulations are provided in Supp. Mat. E).

\noindent\textbf{Automatic Topology Recognition from Grasp.} 
To verify that \acronyme accurately synthesizes the specified contact topologies, we introduce an automated classification step for the generated grasps. First, we extract the subset of gripper points in physical contact with the object (using Eq.~\ref{eq:gripper_contact_def}) and map them back to their semantic zone identifiers (visible in Fig.~\ref{fig:grasp_taxonomy_contact}).
We then compare these observed active zones against the ground-truth required zones of each target template $\calA_m$. We evaluate this match by computing a similarity score $s_m$ based on the Tversky index~\cite{tversky1977features}.
Crucially, we apply an asymmetric penalization that punishes extra, unintended contact regions more strictly than missing ones, effectively penalizing clumsy or degenerated grasps.
Finally, a generated grasp is evaluated against all $M$ templates and assigned the class $m$ yielding the highest similarity score. To ensure robustness and mitigate false positives, any grasp failing to reach a minimum similarity threshold ($s_m < 0.5, \forall m\in[1,M]$) is strictly classified as "\textit{unknown}". (Detailed mathematical formulations of this metric are provided in Supp. Mat. F).

\begin{table}[t]
    \centering
    \resizebox{0.9\textwidth}{!}{
        \begin{threeparttable}
            \begin{tabular}{l|c|c|c|c|ccc}
                \toprule
                \multirow{2}{*}{\textbf{Method}}
                & \multirow{2}{*}{\begin{tabular}[c]{@{}c@{}}\textbf{Object-Agnostic} \\ \textbf{Training} \end{tabular}}
                & \multirow{2}{*}{\textbf{$\textbf{SR}$ $\uparrow$}}
                & \multirow{2}{*}{\textbf{$\textbf{H}_{TC}$ $\uparrow$}}
                & \multirow{2}{*}{\begin{tabular}[c]{@{}c@{}}\textbf{Speed} \\ (sec. / grasps) \end{tabular}}
                & \multicolumn{3}{c}{\textbf{Diversity (avg.) $\uparrow$}}
                \\ 
                \cline{6-8} 
                 &  &  &  &
                 & $\mathbf{t}$ (m) & R (rad) & Q (rad)
                \\ \cline{1-8} 
                DFC~\cite{liu2021synthesizing}
                    & \cmark
                    & \underline{72.15} & \underline{0.7389} &  >1800
                    & \underline{0.0607} & 1.424 & \textbf{0.3579}
                \\
                GenDexGrasp~\cite{li2023gendexgrasp}
                    & \xmark
                    & 71.15 & 0.5956 & 14.65
                    & 0.0519 & 1.416 & 0.2567
                \\
                DRO-Grasp~\cite{wei2024d}
                    & \xmark
                    & 63.30 & 0.6504 & 1.72
                    & 0.0546 & \textbf{1.515} & 0.2892
                \\
                GOAG~\cite{merand2026goag}
                    & \cmark
                    & \textbf{77.90} & 0.6527 & \underline{0.20}
                    & 0.0479 & 1.401 & 0.3170
                \\ \cline{1-8}
                \textbf{\acronyme}
                    & \cmark
                    & 36.94 & \textbf{0.83} & \textbf{0.11}
                    & \textbf{0.0674} & \underline{1.4927} & \underline{0.3458}
                \\
                \bottomrule
            \end{tabular}
        \end{threeparttable}
    }
    \caption{\textbf{Comparison with taxonomy-unaware baselines.} \acronyme achieves the highest semantic entropy ($\textbf{H}_{TC}$) and generation speed, overcoming the functional mode collapse typical of unconditioned planners.}
    \label{tab:tax_unaware_stats}
\end{table}

\subsection{Evaluation Metrics}
\label{subsec:metrics}

To evaluate \acronyme, we utilize a combination of standard physical stability metrics and novel semantic compliance metrics. For physical stability (Success Rate, \textbf{SR}), Generation \textbf{Speed}, and Spatial \textbf{Diversity}, we strictly follow the evaluation protocols established in~\cite{merand2026goag, wei2024d, li2023gendexgrasp} (Detailed formulations for these metrics are provided in Supp. Mat. G). \\

To quantify the functional accuracy and distributional fairness of the generated grasps, we introduce the following novel metrics:

\noindent\textbf{Topology Compliance (TC).} This metric evaluates how accurately the generated grasp adheres to the functional constraints of the requested contact topology. Across the $M$ evaluated topologies, the average \textbf{TC} is defined as:
\begin{equation}
    \textbf{TC} = \frac{1}{M} \sum_{m=1}^{M} \frac{\text{Effective}_{m}}{\text{Attempt}_{m}}
\end{equation}
where $\text{Attempt}_{m}$ is the total number of simulated grasps conditioned on target topology $m$, and $\text{Effective}_{m}$ is the subset of those grasps that successfully satisfy the valid semantic contact template for topology $m$.

\noindent\textbf{Entropy.} To evaluate the topological diversity and assess potential mode collapse, we compute the normalized Shannon entropy across all contact topologies:
\begin{equation}
    \textbf{H} = \frac{-\sum_{m=1}^{M} \tilde{f}_m\ln(\tilde{f}_m)}{\ln(M)}
\end{equation}
A value of $\textbf{H} \to 1$ indicates a uniform, unbiased generative distribution, whereas $\textbf{H} \to 0$ reflects severe mode collapse toward a few dominant topologies.
We report two distinct variants of this metric: \\
\underline{Stability Entropy ($\textbf{H}_{SR}$):} Evaluates if the model generates physically stable grasps equally well across all topologies. Here, $\tilde{f}_m$ is the normalized \textbf{SR} of topology $m$, defined as $\tilde{f}_m = \textbf{SR}_{m} / \sum_{j=1}^M \textbf{SR}_{j}$. \\
\underline{Semantic Entropy ($\textbf{H}_{TC}$):} Evaluates if the model preserves true functional intent without defaulting to simpler power grasps. Here, $\tilde{f}_m$ is the normalized \textbf{TC} of the topology $m$, defined as $\tilde{f}_m = \textbf{TC}_{m} / \sum_{j=1}^M \textbf{TC}_{j}$.

\subsection{Overall Performances}
\label{subsec:overall_performances}

\begin{figure}[t]
    \centering
    \includegraphics[height=0.25\textheight, width=\columnwidth]{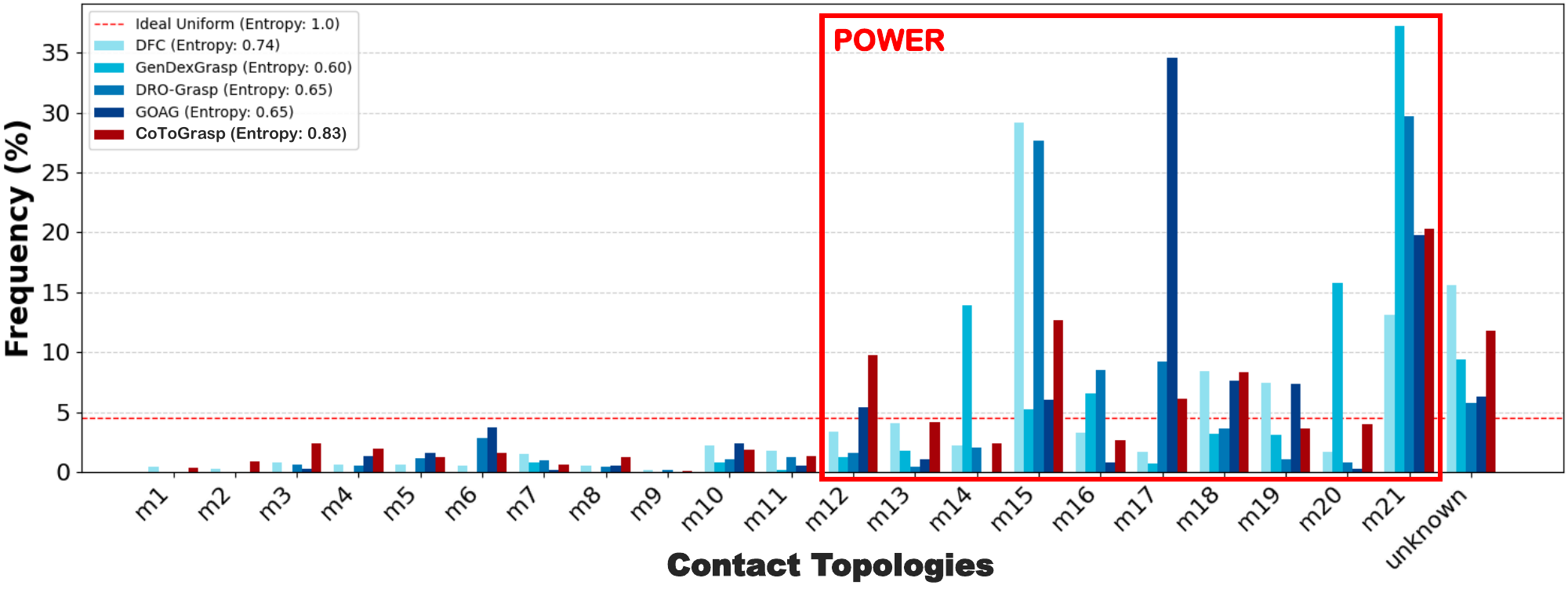}
    \caption{\textbf{Functional contact topology distribution across taxonomy-unaware planners.} The histogram illustrates the distribution of grasps generated by unconditioned baselines compared to \acronyme on the Multidex objects set. 
    The unknown category represents physically stable grasps with unnatural contact patterns that fail to match any contact topology.
    Notably, unconditioned baselines exhibit a severe generative bias (mode collapse) toward enveloping power grasps (red box).}
    \label{fig:type_diversity}
\end{figure}

We evaluate \acronyme against state-of-the-art baselines using the Shadow Hand. First, we evaluate against unconditioned generative planners to highlight the necessity of topology conditioning. Second, we evaluate against a taxonomy-aware baseline to demonstrate \acronyme's superior functional control. 
For all experiments, we cap generation at 20 attempts per topology. This resampling budget mitigates heuristic approach poses $(R, \mathbf{t})$ that are geometrically incompatible with the requested contact topology.

\noindent\textbf{Taxonomy-Unaware Grasp Planners Analysis and Comparison.}
Evaluating on a MultiDex subset, Table~\ref{tab:tax_unaware_stats} shows unconditioned baselines~\cite{liu2021synthesizing,li2023gendexgrasp,wei2024d,merand2026goag} achieve higher success rates (SR) by naturally defaulting to geometrically stable, enveloping power grasps (Fig.~\ref{fig:type_diversity}). Consequently, they struggle with precision grasps, yielding low semantic entropy ($\textbf{H}_{TC}$). Conversely, \acronyme achieves a balanced topology distribution and the highest $\textbf{H}_{TC}$. Its overall SR (36.94\%, with a Top-5 average of 56.18\%) occurs because it is forced to synthesize diverse topologies (\eg, precision pinches) regardless of an object's natural geometric affordances, inherently prioritizing target contacts over simulated physics stability. While methods like DFC show high spatial variance ($\mathbf{t}, R, Q$), this primarily reflects joint range exploitation rather than functional diversity. Furthermore, \acronyme demonstrates the fastest generation speed at $0.11$s per grasp.

\begin{table}[tp]
    \resizebox{\textwidth}{!}{
    \begin{tabular}{l|ccc|c|c|c|c|c}
        \toprule
        \multirow{2}{*}{\textbf{Method}}
        & \multicolumn{5}{c|}{\textbf{SR (\%)}}
        & \multirow{2}{*}{$\textbf{H}_{SR}$}
        & \multirow{2}{*}{\textbf{TC (\%)}}
         & \multirow{2}{*}{$\textbf{H}_{TC}$}
        \\ \cline{2-6}
         & Power & Precision & Obj. Spe. & Avg. Topo. & Avg. Obj. & & &
        \\ \midrule
        Dexonomy~\cite{chen2025dexonomy} & 27.16 & 12.36 & 19.62 & 21.13 & 23.80 & 0.91 & 14.28 & 0.77
        \\
        \textbf{\acronyme} & \textbf{29.75} & \textbf{22.71} & \textbf{25.50} & \textbf{26.72} & \textbf{27.56} & \textbf{0.96} & \textbf{17.18} & \textbf{0.84}
        \\ \midrule
        \acronyme (w/o Label-Consistency) & 25.11 & 14.77 & 20.85 & 21.14 & 22.97 & 0.94 & 14.45 & 0.81
        \\
        \acronyme (w/o Force-Closure) & 26.90 & 14.87 & 21.08 & 22.08 & 23.65 & 0.95 & 16.26 & 0.81
        \\
        \acronyme (No Check) & 25.09 & 14.73 & 20.58 & 21.06 & 23.00 & 0.94 & 14.72 & 0.81
        \\
        \bottomrule
    \end{tabular}
    }
    \caption{\textbf{Taxonomy-aware grasp synthesis.} \acronyme outperforms the baseline in both physical stability (SR) and semantic accuracy (TC), particularly on highly constrained precision grasps.}
    \label{tab:tax_aware_results}
    \vspace{-20pt}
\end{table}

\noindent\textbf{Taxonomy-Aware Grasp Synthesis.}
To evaluate precise functional control, we compare \acronyme against Dexonomy~\cite{chen2025dexonomy} on the DexGraspNet~\cite{wang2023dexgraspnet} test set. After mapping Dexonomy's Feix~\cite{feix2015grasp} taxonomy to our contact-based representations for direct comparison (Fig.~\ref{fig:grasp_taxonomy_contact}), Table~\ref{tab:tax_aware_results} shows \acronyme achieves higher \textbf{SR} and \textbf{TC} ($17.18\%$ vs. $14.28\%$) across all functional categories. While absolute \textbf{TC} scores appear modest, they reflect the extreme stringency of our asymmetric metric, which heavily penalizes the minor, physically necessary finger adjustments naturally occurring during dynamic simulation. Outperforming the baseline under these strict conditions demonstrates \acronyme's superior zero-shot functional control.
Dexonomy~\cite{chen2025dexonomy} struggles with \textbf{TC} because its conditioning is rigidly coupled to explicit joint configurations end-to-end; when object geometry forces deviations, the resulting grasp frequently violates the requested topology. Conversely, \acronyme robustly projects valid semantic contact zones via object-agnostic training. The ablation of our validation pipeline (No Check) demonstrates a sharp performance drop, emphasizing that verifying topological viability against the object's local geometry via the Label-Consistency check is critical before optimization. (A detailed analysis of \textbf{TC} bottlenecks before and after physics simulation is provided in Supp. Mat. I).

\begin{wraptable}{r}{0.50\textwidth}
    \centering
    \vspace{-15pt}
    \resizebox{\linewidth}{!}{
    \begin{tabular}{l|cccccc}
        \toprule
        \multirow{2}{*}{\textbf{Method}}
        & \multicolumn{6}{c}{\textbf{SR (\%)}}
        \\ \cline{2-7}
         & \textbf{M2} & \textbf{M6} & \textbf{M11} & \textbf{M13} & \textbf{M18} & \textbf{M21}
        \\  \midrule
        Dexonomy~\cite{chen2025dexonomy} & 10.5 & 15.2 & 60.5$^\dagger$ & 20.3 & 29.6 & 37.2$^\dagger$
        \\
        \textbf{\acronyme} & \textbf{30.3} & \textbf{21.7} & 29.8 & \textbf{29.6} & \textbf{31.3} & 33.5
        \\
        \bottomrule
    \end{tabular}
    }
    \vspace{-3pt}
    \caption{Per-Topology results: 3 Power grasps (M13, M18, M21), 2 Precision (M2, M6) and 1 Object-Specific (M11). $^\dagger$Indicates artificially inflated scores due to mode collapse, where Dexonomy defaults to unverified enveloping grasps.}
    \label{tab:results_per_type}
    \vspace{-20pt}
\end{wraptable}

\noindent\textbf{Per-Topology Analysis.}
Table~\ref{tab:results_per_type} highlights \acronyme's distinct advantage in synthesizing highly constrained precision grasps (\eg M2: $30.3\%$ vs. $10.5\%$). 
An apparent pseudo-anomaly occurs with topologies M11 (Object-Specific) and M21 (Pow\-er), where Dexonomy~\cite{chen2025dexonomy} reports artificially inflated \textbf{SR} ($60.5\%$ and $37.2\%$). 
Qualitative analysis reveals this is driven by severe mode collapse: when faced with challenging geometries, Dexonomy~\cite{chen2025dexonomy} frequently abandons the conditional query and defaults to an unverified M21 enveloping grasp. Because it lacks a strict posterior topological check, it erroneously records these power grasps as successes for different input topologies (like M11). Conversely, \acronyme explicitly detects and rejects hallucinated contacts, maintaining a competitive true-positive rate for M21 ($33.5\%$) without sacrificing the integrity of the true \textbf{TC} metric on precision tasks.

\subsection{Real-World Validation}
\label{subsec:real_exp}

\begin{figure}[t]
    \centering
    \includegraphics[width=\columnwidth]{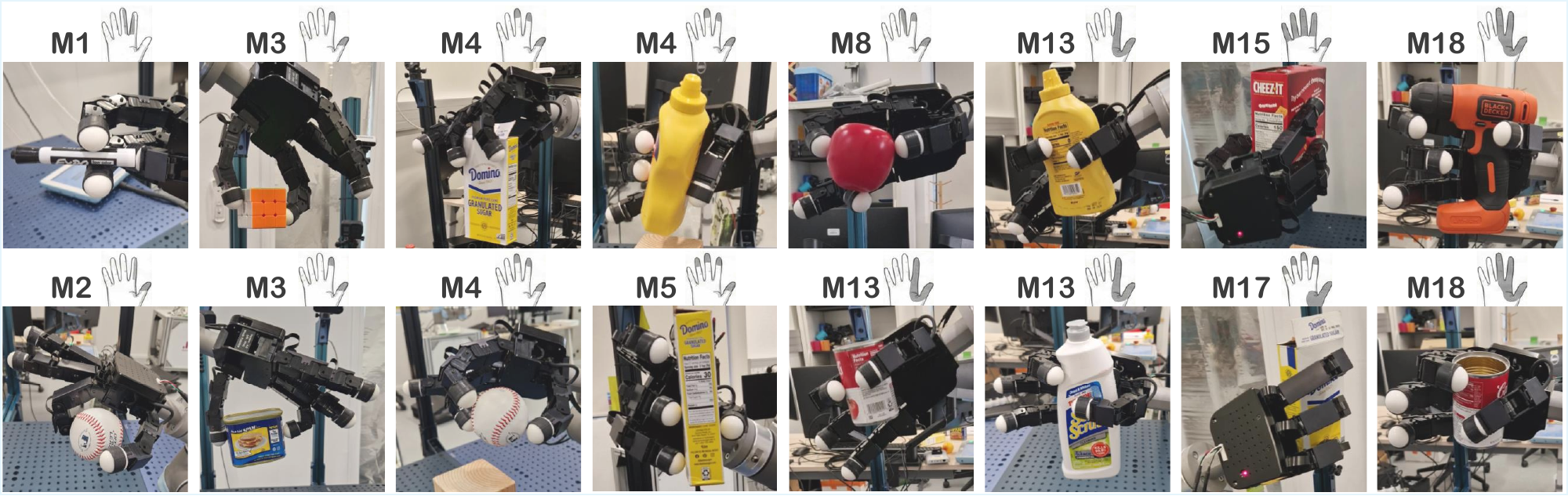}
    \caption{\textbf{Real-World kinematic validation.} \acronyme synthesizes diverse, topology-compliant grasps that are physically executable on a physical Allegro Hand using YCB~\cite{calli2015ycb} objects. The target contact topologies (indicated above each frame) demonstrate the physical viability of the generated grasps across both precision and power categories.}
    \label{fig:real_exp}
\end{figure}

To demonstrate the physical viability and kinematic feasibility of the grasps synthesized by \acronyme, we conducted real-world validation experiments. These were performed in a tabletop environment using a 6-DoF Universal Robots UR10 manipulator equipped with an Allegro Right Hand. Because the Allegro Hand is a four-fingered anthropomorphic gripper (lacking a little finger), we explicitly suppressed the generation of contact topology M6, as it strictly requires five fingers to form a valid contact template.
We successfully planned and executed grasps on diverse objects from the YCB~\cite{calli2015ycb} dataset. Crucially, we empirically validated the functional diversity of our method by demonstrating that the generated grasp for various contact topologies can be successfully formed and held statically on the physical system. A qualitative overview of the successful real-world grasps, categorized by their respective contact topologies, is presented in Figure~\ref{fig:real_exp}.
Execution sequences are provided in the supplementary video material. The inherent hardware challenges and control limitations of executing precise semantic grasps using a multifingered gripper like the Allegro Hand are discussed extensively in Supp. Mat. J.
\section{Conclusion}

This paper introduces \acronyme, a novel generative framework for contact-topology-conditioned dexterous grasp synthesis. By fundamentally decoupling functional intent from object identity, our approach effectively mitigates the severe mode collapse observed in contemporary grasp planners. This allows \acronyme to synthesize highly constrained, topology-compliant precision and power grasps on unseen geometries without relying on costly object-annotated datasets. Extensive evaluations demonstrate that our validation pipeline robustly filters topologically invalid configurations, yielding state-of-the-art semantic diversity and physical stability. Finally, successful real-world deployments on the Allegro Hand confirm that the structural properties of our generated contact topologies are physically executable on a real robot platform.

\section*{Acknowledgments}

This publication was made possible by the use of the FactoryIA supercomputer, financially supported by the Ile-De-France Regional Council.
Experiments presented in this paper were carried out thanks to a platform funded by DIM AI4IDF and PRAIRIE-PSAI. The authors would like to thank Timothée Carecchio for his valuable assistance in achieving the experimental results presented in this paper.
This project has received funding from the European Union’s Horizon Europe research and innovation program under grant agreement nº 101135708 (JARVIS Project).  Liming Chen in this research  was in part supported by the French Research Agency ANR, l’Agence Nationale de Recherche, through the projects Aristotle (ANR-21-FAI1-0009-01),  Astérix (ANR-23-EDIA-0002), DEMETER  (ANR-25-HTCE-0002) and PROTEUS (ANR-25-TSIA-0011-01), and the French national investment prioritary program through the PSPC FAIR WASTE project.

%
%
\bibliographystyle{splncs04}
\bibliography{main}

\clearpage
\appendix

\begin{center}
   \Large \bfseries Supplementary Material
\end{center}

This supplementary material provides additional technical details to support the main manuscript. Section~\ref{supp_mat:sec:tax_transfer_handprint} details the taxonomy transfer and handprint discretization. Section~\ref{supp_mat:sec:architecture} outlines the network architecture and hyperparameters. Section~\ref{supp_mat:sec:domain_shift} justifies the train-test domain shift quantitatively and qualitatively. Section~\ref{supp_mat:sec:validation_pipeline} provides the mathematical formulations for our validation pipeline and energy-based optimization. Finally, Sections~\ref{supp_mat:sec:pose_sampling} through~\ref{supp_mat:sec:grasp_visu} offer extended experimental protocols, metric definitions, and comprehensive qualitative results.

\section{Applying contact topologies to grippers}
\label{supp_mat:sec:tax_transfer_handprint}

\begin{figure}[!htbp]
    \centering
    \includegraphics[width=0.6\textwidth]{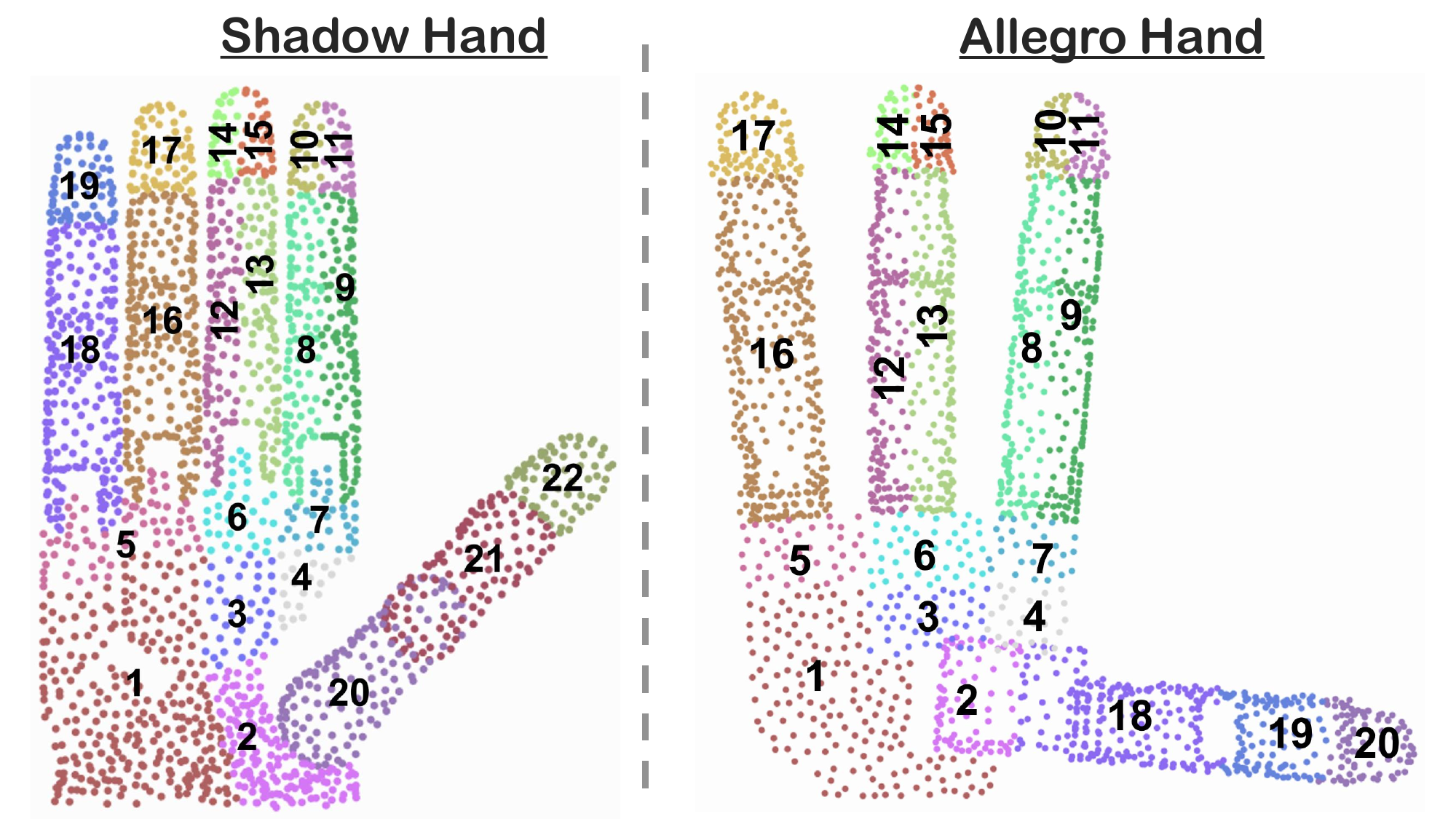}
    \hfill
    \includegraphics[width=0.3\textwidth]{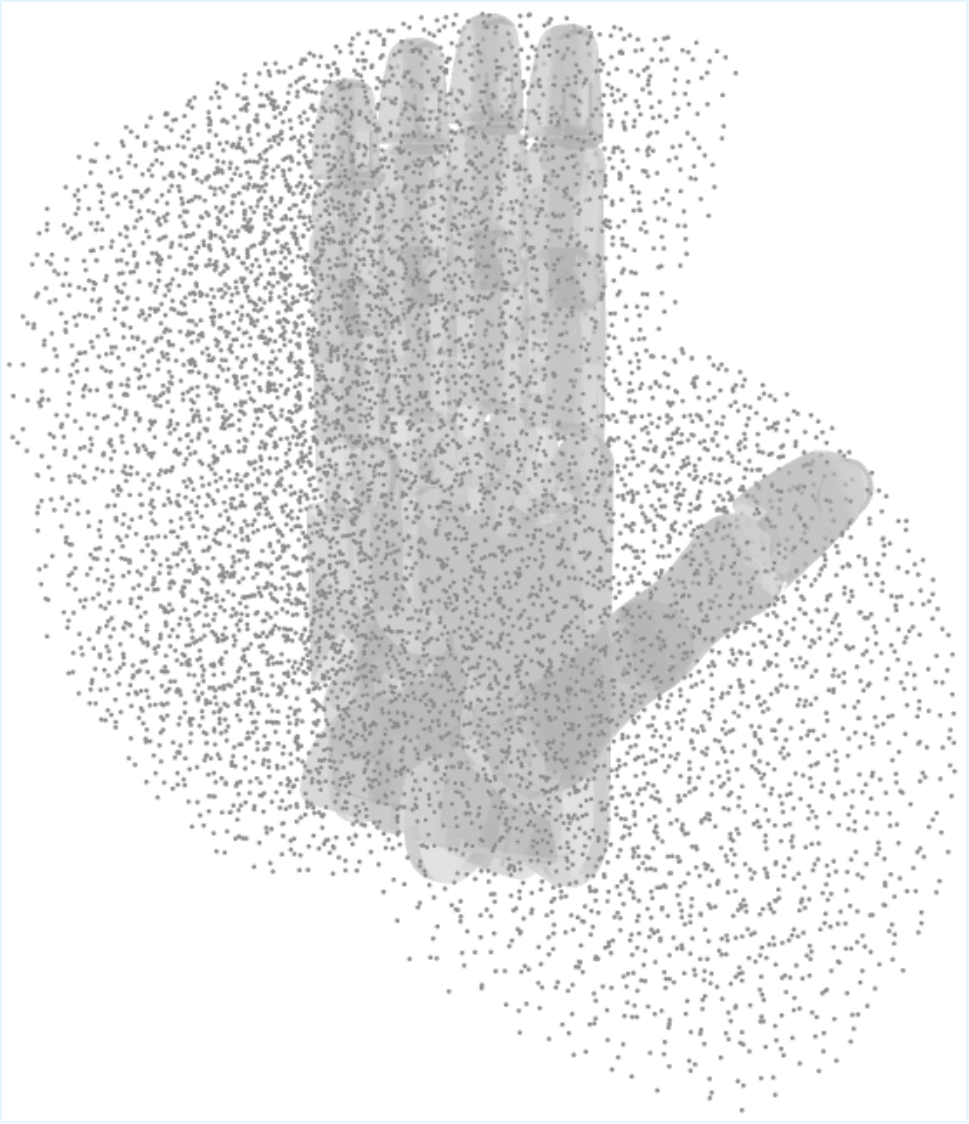}
    \caption{\textbf{Handprint areas  and workspace constitution.} \textbf{Left:} Discretized handprints of the Shadow and Allegro hands, illustrating the manually defined anatomical zone divisions. \textbf{Right:} A three-quarter view of the Shadow Hand's workspace.}
    \label{supp_mat:fig:grippers_zones}
\end{figure}

\noindent\textbf{Handprint Discretization.}
To generate the spatial handprints $\calH(Q)$ for the training dataset, we compute the forward kinematics for each sampled joint configuration $Q$. To represent the gripper's active grasping surfaces, we discretize its geometry into a fixed point cloud of $N_H = 2,048$ points. We filter the dense surface points based on their spatial dot products relative to the approach direction, ensuring a uniform, configuration-independent resolution across the functional contact areas.
Specifically, let $\mathbf{G} \in \mathbb{R}^{M \times 3}$ represent the matrix of surface normals for a dense set of $M$ candidate points on the gripper's geometry. We define a reference direction vector, $\mathbf{v}_{\text{ref}} \in \mathbb{R}^{3 \times 1}$, representing the palm's facing direction (\eg $ \mathbf{v}_{\text{ref}} = [0, -1, 0]^T$ for the Shadow Hand). To isolate the active grasping surfaces, we evaluate the alignment of all candidate points simultaneously by computing the dot products between their normals and the reference direction:
\begin{equation}
    \mathbf{s} = \mathbf{G} \mathbf{v}_{\text{ref}}
\end{equation}
The resulting scalar values in the vector $\mathbf{s} \in \mathbb{R}^{M \times 1}$ are then used to threshold and sample the most relevant contact surfaces, culminating in the final $N_H$ points that form $\calH(Q)$.
Figure~\ref{supp_mat:fig:grippers_zones} shows the resulting handprints for the Shadow and Allegro hands, alongside the manually defined zone divisions detailed in Section~\ref{subsec:taxonomy_formulation}.

\begin{figure}[!htbp]
    \centering
    \includegraphics[width=\textwidth]{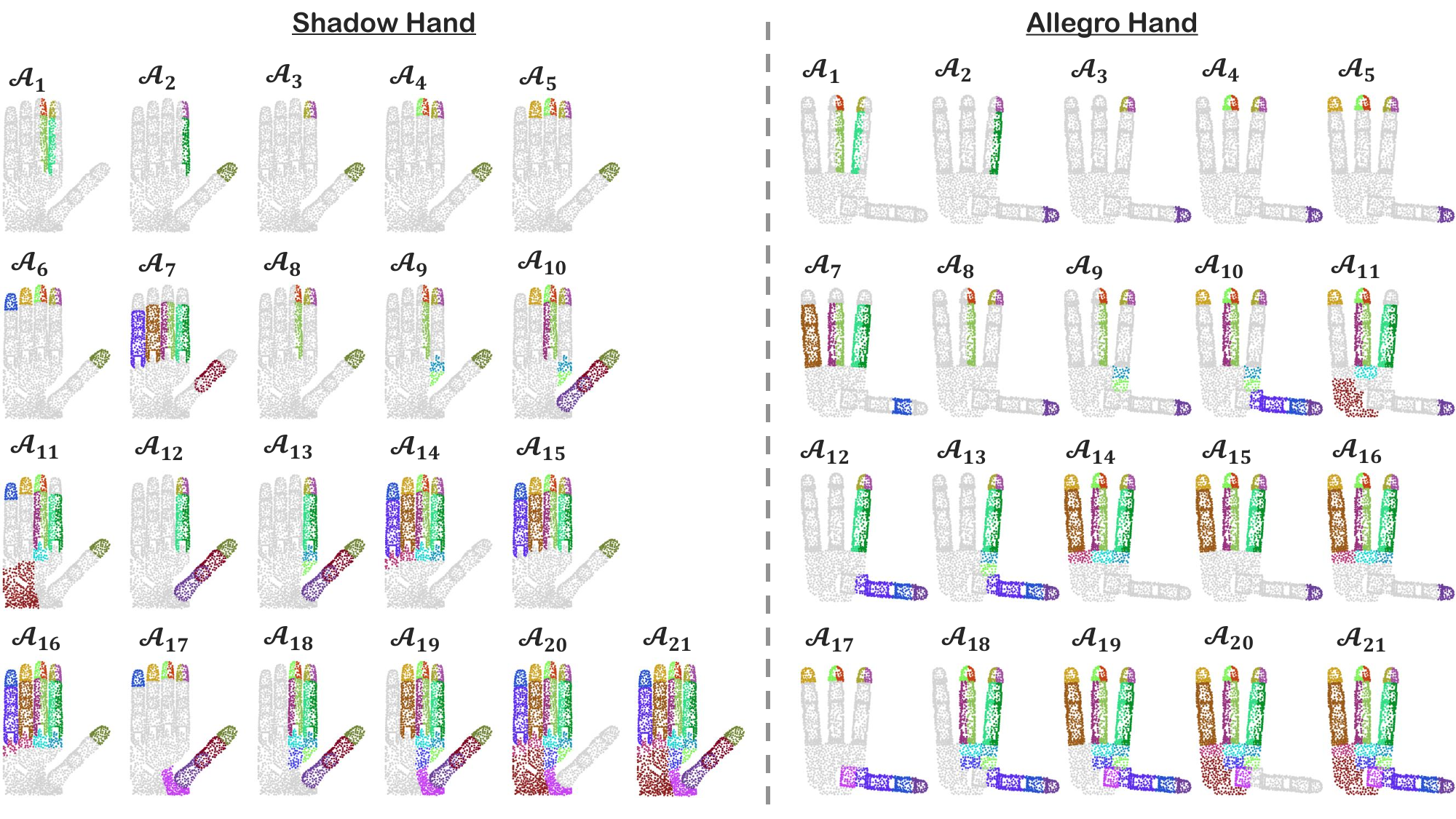}
    \caption{\textbf{Taxonomy transfer mapping for anthropomorphic grippers.} The handprints of the Shadow Hand (left) and the Allegro Hand (right) are segmented into manually defined, corresponding anatomical zones ($\mathcal{A}_1$--$\mathcal{A}_{21}$).}
    \label{supp_mat:fig:tax_transfer}
\end{figure}

\noindent\textbf{Taxonomy Transfer to Anthropomorphic Grippers.}
To semantically ground the generated handprints $\mathcal{H}(Q)$, we map established human grasp taxonomies onto the robotic kinematics. We manually partition the active grasping surfaces of both the Shadow Hand and the Allegro Hand into a discrete set of anatomical zones. Specifically, the mechanical links and palm areas are segmented into the $M=21$ contact topology templates. During the handprint generation process, each sampled point $\mathbf{h} \in \calH(Q)$ is assigned to its corresponding zone $\calA_k$. Notably, because the Allegro Hand lacks a fifth finger, the contact topology templates for $\calA_5$ and $\calA_6$ are functionally identical. To prevent redundancy, we explicitly omit $\calA_6$ from the Allegro Hand's template set.
Ultimately, this taxonomy transfer (illustrated Figure~\ref{supp_mat:fig:tax_transfer}) ensures that despite the morphological differences between the grippers, their spatial handprints share a common semantic representation, enabling structured, cross-embodiment comparisons of grasp strategies. 

\noindent\textbf{Workspace Computation.}
We define the gripper's workspace as an aggregate point cloud representing its reachable surface area. This representation is generated by superimposing 10,000 gripper handprints sampled from our synthetic dataset, effectively capturing the complete interaction space of the fingers and palm independently of any external environment. To construct our fixed-size set of spatial basis points $\calW = \{w_j \in \mathbb{R}^{3}\}_{j=1}^{N_W}$, this dense accumulation is downsampled using Farthest Point Sampling (FPS). This yields a spatially uniform distribution of $N_W$ that comprehensively encapsulate the gripper's kinematic workspace.
Figure~\ref{supp_mat:fig:grippers_zones} shows an illustration of the Shadow Hand's workspace.

\section{Architectural Choices and Implementation Details}
\label{supp_mat:sec:architecture}

To ensure the \acronyme framework accurately models the complex physics and semantics of dexterous grasping, several deliberate architectural choices were implemented. This section details the theoretical justifications for these design choices.

\noindent\textbf{Local Geometry Extraction vs. Global Shape Descriptors.}
Standard 3D generative pipelines often employ global shape descriptors (\eg standard PointNet~\citesupp{qi2017pointnet}) that compress the entire object into a single latent vector. While effective for classification, global compression destroys high-frequency spatial details -- such as local curvature and surface normals -- which are absolutely critical for identifying physically viable contact patches. 
For this, we implement a DGCNN~\cite{phan2018dgcnn} encoder using the architecture presented in \cite{wei2024d}. Unlike the standard \textit{Edge Convolution} operation that dynamically recomputes $K$-Nearest Neighbors at every network layer, this architecture utilizes a fixed graph structure throughout, resulting in a "Static Graph CNN". Specifically, we set the neighborhood size to $K = 16$ to capture highly localized geometric information. \\
By utilizing this "Static Graph CNN" architecture, we preserve point-wise local features $\calF_{\calP}$, ensuring the network reasons over local geometric fidelity while remaining invariant to the global spatial arrangement of the point cloud. Its learned weights successfully generalize across the domain shift -- from the structured kinematic handprint seen during training to the highly variable target objects processed during inference.

\noindent\textbf{Features Aggregation during the Workspace Projection.}
During the feature projection phase onto the canonical workspace $\calW$, we utilize a modified $k$-Nearest Neighbors aggregation. Inspired by scattered data interpolation in neural rendering~\citesupp{xu2022point}, this local pooling strategy preserves high-frequency details more effectively than dense voxel grids or global pooling. In standard scattered data interpolation, it is common to normalize the aggregated features by the sum of the exponential weights. 
We intentionally omit this normalization step, instead computing an unnormalized average divided strictly by $k$ (Eq.~\ref{eq:features_projection}). Because the workspace covers the entire kinematic reach of the gripper, many basis points lie in "empty space," far from the input geometry. By leaving the distance weights unnormalized, they naturally decay to zero for distant neighbors. This consequently suppresses feature activation in empty regions, effectively preventing the network from hallucinating object geometry where none exists.

\noindent\textbf{Persistent Hard Conditioning in Self-Attention.}
Standard vision transformer architectures~\citesupp{dosovitskiy2021image} (ViT) typically condition the generation process by prepending the condition variable as an isolated global [CLS] token. However, grasp synthesis relies heavily on precise local contact arrangements across thousands of spatial tokens. 
A single global token is prone to signal dilution across deep attention layers~\citesupp{zhou2021deepvit}.
To circumvent this, we explicitly concatenate the semantic topology embedding $\calF_{\calT}$ to every single workspace point feature. This point-wise fusion enforces a persistent, "hard" conditioning across the entire spatial domain, ensuring that every local geometric feature is evaluated strictly under the lens of the target functional intent at every layer of the network. \\
Since the projected workspace features $\calF_{\calW}$ are processed by the transformer as an unordered sequence of tokens -- lacking the explicit 3D coordinates $w_i$ of the original basis points -- we inject spatial awareness via sinusoidal positional encodings (PE). This allows the multi-head attention mechanism to recover spatial relationships purely from the fixed basis set indices.
Crucially, the network is not provided with explicit contact priors at inference time. Instead, the semantic knowledge of valid contact configurations is acquired implicitly through backpropagation. The supervision signal drives the multi-head self-attention mechanism to discover and amplify geometric correlations between spatially distant points that correspond to stable configurations.

\noindent\textbf{Global Latent Pooling via Set Transformer.}
Because our architecture explicitly omits a global learnable [CLS] token to preserve dense local conditioning, we cannot simply extract a designated output token (such as the $\mathbf{z}_{L}^{0}$ state in standard ViTs) to form our global representation. Consequently, an explicit aggregation strategy is required to synthesize the dense sequence of fused geometric and semantic features into a unified global latent descriptor $\calZ$. For this task, we opted against static pooling operators (\eg max-pooling or average-pooling). Static operators process elements independently, discarding non-extremal data and failing to capture spatial correlations. Instead, we utilize a Set Transformer. The Pooling by MultiHead Attention (PMA) block adaptively weighs input instances based on task relevance, while the subsequent Set Attention Block (SAB) explicitly models pairwise and higher-order interactions. This allows the network to successfully encode the structural co-occurrences and spatial distributions of distant contact patches.

\noindent\textbf{CVAE and Contact-Topology-Conditioned Multi-Modality.}
A well-doc\-umented limitation of standard CVAEs in grasp generation is their tendency to suffer from posterior collapse~\cite{huang2023diffusion}, \citesupp{fu2019cyclical}, \citesupp{he2019lagging} or blurred predictions when faced with the highly multi-modal nature of general grasping (\eg the exact same object can be pinched, enveloped, or grasped by the rim). \acronyme bypasses this limitation. Because the macroscopic multi-modality is explicitly resolved by the strict topology conditioning (which dictates the exact functional approach), the CVAE is relieved of the burden of modeling drastically different grasp modes. It is only required to model the localized, intra-topology spatial variations of the contact patches (\eg slight finger shifts along an edge). For this highly constrained stochasticity, a standard unimodal Gaussian prior $\calN(0,I)$ proves to be both mathematically sufficient and computationally highly efficient.
The variational lower bound (ELBO) consists of the reconstruction term, a Multi-Class Cross-Entropy loss ($L_{\text{recon}}$) over the workspace points $\calJ = \{1, \dots, n_W\}$, and a Kullback-Leibler regularization term:
\begin{equation}
    L = L_{\text{recon}} + \beta D_{\text{KL}}(q(z|\calZ) ~\|~ \calN(0,I)) 
    \text{,} \quad 
    L_{\text{recon}} = - \sum_{j \in \calJ} \mathbf{y}_j \cdot \log(\hat{\mathbf{y}}_j)
\end{equation}
where $\beta$ is a weighting parameter balancing latent capacity and reconstruction accuracy, $\mathbf{y}_j$ is the one-hot encoded ground truth derived from $\Lambda_m(j)$ and $\hat{\mathbf{y}}_j$ is the predicted probability vector.

\noindent\textbf{Network Hyperparameters and Training Details.} 
Table \ref{supp_mat:tab:hyperparameters} summarizes the comprehensive architectural details, layer dimensions, and hyperparameters used to implement and train the \acronyme framework. All training procedures were executed across a cluster of 4 Nvidia A100 GPUs, requiring approximately 35 hours.

\begin{table}[!t]
    \centering
    \resizebox{0.45\textheight}{!}{
    \begin{tabular}{llc}
        \toprule
        \textbf{Module / Parameter} & \textbf{Notation} & \textbf{Value / Size} \\
        \midrule
        \multicolumn{3}{c}{\textbf{Training \& Optimization}} \\
        \midrule
        Batch Size & -- & 32 \\
        Learning Rate & -- & $10^{-5}$ \\
        Training Epochs & -- & 50 \\
        KLD Regularization Weight & $\beta$ & $0.1$ \\
        Attention Factor & $\alpha$ & $3.0$ \\
        \midrule
        \multicolumn{3}{c}{\textbf{Workspace \& Geometric Projection}} \\
        \midrule
        Workspace Resolution & $N_W$ & $8,192$ \\
        Projection kNN & $k$ & $5$ \\
        Aligned Distance Scaling & $\gamma$ & $2.0$ \\
        \midrule
        \multicolumn{3}{c}{\textbf{Pointwise Feature Extraction}} \\
        \midrule
        Local Graph kNN & $K$ & $16$ \\
        Hidden Layer Sizes & -- & $[12, 64, 64, 128, 256, 512]$ \\
        Output Feature Dimension & $\calF_{\calP}, \calF_{\calW}$ & $1,024$ \\
        \midrule
        \multicolumn{3}{c}{\textbf{Conditioning Embeddings}} \\
        \midrule
        Topology Embedding Dim. & $\calF_{\calT}$ & $64$ \\
        Label Embedding (MLP) & $\calF_{\Lambda}$ & $[128, 64]$ \\
        \midrule
        \multicolumn{3}{c}{\textbf{Self-Attention Modules}} \\
        \midrule
        Transformer Encoder Feature Dim. & $\Phi$ & 1088 \\
        Transformer Encoder Blocks & -- & $4$ \\
        Transformer Encoder Heads & -- & $8$ \\
        Set Transformer (Pooling) Heads & -- & $8$ \\
        Global Latent Descriptor Dim. & $\calZ$ & $1,024$ \\
        \midrule
        \multicolumn{3}{c}{\textbf{CVAE \& Latent Space}} \\
        \midrule
        Latent Encoder Hidden Dim. & -- & $512$ \\
        Latent Variable Dimension & $\psi$ & $64$ \\
        AdaLN Decoder Output Classes & $N + 1$ & $23$ \\
        \bottomrule
    \end{tabular}
    }
    \caption{Implementation Details and Network Hyperparameters for the \acronyme framework.}
    \label{supp_mat:tab:hyperparameters}
    \vspace{-10pt}
\end{table}

\section{Train-Test Domain Shift and Feature Alignment}
\label{supp_mat:sec:domain_shift}

To demonstrate that our architectural choices enable robust cross-domain generalization -- specifically, the transfer from the gripper surface (training domain) to the object surface (inference domain) -- we analyze the latent feature alignment across both domains. This evaluation verifies whether our framework successfully bridges the inherent geometric and structural differences between hands and objects.

\noindent\textbf{Quantitative Alignment Analysis.} 
To measure the transfer quality, we extracted intermediate feature representations from both domains and evaluated their alignment at active topological contact points using Average Cosine Similarity. We compared corresponding spatial points (\textit{Matched Pairs}) against random, non-corresponding points (\textit{Random Pairs}) to establish a baseline (Table~\ref{supp_mat:tab:feature_alignment}).
\begin{table}[htbp]
    \centering
    \resizebox{0.6\textwidth}{!}{
    \begin{tabular}{lccc}
        \hline
         & Raw DGCNN & Workspace & + Attn. \\ \hline
        Matched Pairs & \textbf{0.35} & \textbf{0.61} & \textbf{0.66} \\ \hline
        Random Pairs & 0.23 & 0.33 & 0.27 \\ \hline
    \end{tabular}
    }
    \caption{Feature Alignment (Cosine Similarity).}
    \label{supp_mat:tab:feature_alignment}
\end{table}
Initially, the raw point-wise features extracted from the DGCNN ($\calF_\calP$) exhibit a significant domain gap, achieving a poor similarity score of $0.35$ for matched pairs and $0.23$ for random pairs. This confirms that the raw domain shift remains substantial despite static graph modifications. However, the effectiveness of projecting into our feature-based canonical workspace via kNN aggregation ($\calF_\calW$) is immediately evident: the matched pair cosine similarity jumps to $0.61$. This quantitatively proves that the canonical projection successfully disentangles features from the input topology, effectively bridging the domain gap. 
Finally, introducing the contact topology embedding ($\calF_\calT$) into the Transformer encoder ($\Phi$) further sharpens this representation. The self-attention mechanism increases matched similarity to $0.66$ while actively suppressing mismatched, domain-specific geometric noise (reducing random pair similarity from $0.33$ to $0.27$). 

\begin{figure}[htbp]
    \centering
    \includegraphics[width=0.8\columnwidth]{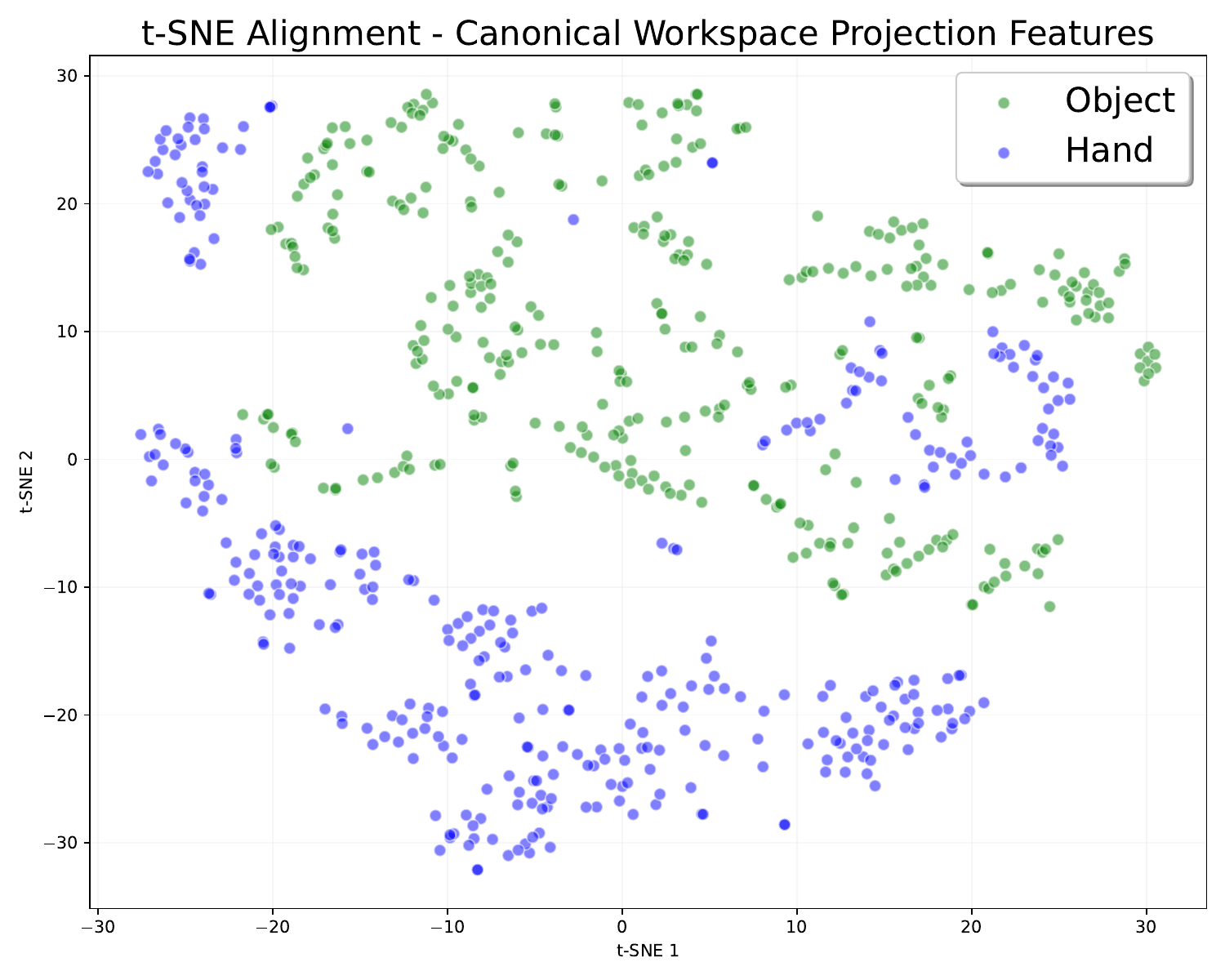}
    \caption{t-SNE visualization of features after canonical workspace projection.}
    \label{supp_mat:fig:domain_shift}
\end{figure}

\noindent\textbf{Qualitative Latent Space Visualization.} 
This quantitative improvement is visually corroborated by the t-SNE projection of our canonical workspace features ($\calF_\calW$), illustrated in Figure~\ref{supp_mat:fig:domain_shift}. 
While global domain distinctions naturally persist -- reflecting the inherent macro-structural differences between an anthropomorphic hand and arbitrary objects -- the features distinctly interleave within shared manifold clusters at task-relevant regions. 
This selective alignment visually and mathematically justifies our reliance on the self-attention layers ($\Phi$). The Transformer is necessary to isolate and attend to these domain-invariant contact primitives, resolving non-local spatial dependencies while ignoring irrelevant structural disparities. Ultimately, this ensures that the generative process remains robust regardless of the input surface.

\section{Validation Pipeline and Energy Optimization Details}
\label{supp_mat:sec:validation_pipeline}

This section details the post-inference validation and optimization pipeline, which is designed to filter out physically and topologically invalid predictions before finalizing the hand's kinematics. To ensure both semantic correctness and physical feasibility, our framework processes the generated contact templates through a sequential validation strategy: a label-consistency check, a preliminary force-closure evaluation, and an energy-based joint optimization. The average runtime per grasp evaluates to 110 ms, with the computational cost broken down as follows: pose initialization (2.8 ms), forward network inference (0.3 ms), label-consistency verification (42.4 ms), force-closure computation (14.7 ms), and the final joint configuration optimization (50.7 ms).

\noindent\textbf{Label-Consistency Check Formulation.}
Let $\calU(\Lambda) = \{ \Lambda(j) \mid \Lambda(j) \neq 0 \}$ be an operator that extracts the set of unique, active semantic zone identifiers from a given contact template. To prevent the generation of ill-posed grasps, we define the validation function $V(\Lambda_m, \hat{\Lambda}_m)$ as a strict binary check on the symmetric difference between the extracted zone sets of the ground truth and predicted templates::
\begin{equation}
\label{eq:label_checker}
    V(\Lambda_m, \hat{\Lambda}_m) = \mathbb{I}(|\calU(\Lambda_m) \Delta \calU(\hat{\Lambda}_m)| \leq 1)
\end{equation}
where $\mathbb{I}$ is the indicator function and $\Delta$ is the symmetric difference operator defined as $\calU(\Lambda_m) \Delta \calU(\hat{\Lambda}_m) = (\calU(\Lambda_m) \setminus \calU(\hat{\Lambda}_m)) \cup (\calU(\hat{\Lambda}_m) \setminus \calU(\Lambda_m))$. Setting the tolerance threshold to 1 allows for a maximum deviation of one missing or hallucinated contact zone (\eg a missing palm contact in a power grasp) while successfully rejecting fundamentally distinct topological failures.

\noindent\textbf{Force-Closure Formulation.}
Unlike methods trained on pre-validated grasp datasets, our generative approach requires an explicit assessment of grasp stability. To ensure the inferred contact points can theoretically yield a stable grasp before performing the computationally expensive joint configuration optimization, we evaluate the Force-Closure condition using the explicit spatial coordinates of the active workspace points.
\begin{itemize}
    \item \textbf{Contact Modeling:} We first extract the subset of workspace points assigned a valid contact label: $\hat{\calC}(\calW) = \{w_j \in \calW \mid \hat{\Lambda}_m(j) \neq 0\}$. We then group these active points by their predicted semantic zone indices. To enforce a unique contact location per required zone, we compute the spatial barycenter of each point cluster and project it onto the object’s surface $\tilde{\calO}$, yielding a set of discrete contact locations $b_i$. We assume a Coulomb friction model with a friction coefficient of $\mu = 0.3$. Because this estimation is performed entirely within the canonical gripper's reference frame to assess geometric feasibility, we do not factor in gravity or the object's mass at this stage.
    \item \textbf{Force-Closure Computation:} We compute the grasp wrench space, defined as the convex hull of all possible wrenches generated by unit contact forces at locations $b_i$. The total wrench is given by:$$d = \sum_{i=1}^{k} d_i = \sum_{i=1}^{k} G_i f_i$$where $G_i$ is the partial grasp matrix for contact $b_i$, and $f_i$ represents the primitive force vectors along the edges of the friction cone, normalized such that $\|f_i\| = 1$. We verify the force-closure condition by checking if the origin of the wrench space lies strictly within the interior of this convex hull.
    \item \textbf{Theoretical vs. Physical Quality:} It is important to note that this analytical step strictly validates the theoretical stability potential of the semantic contact configuration using a simplified barycenter model. The final, true physical grasp quality is determined during the energy-based optimization, which forces the gripper's complex kinematics to align with the full, dense 3D points of $\hat{\calC}(\calW)$ rather than just these idealized discrete barycenters.
\end{itemize}

\noindent\textbf{Joint Configuration Optimization Formulation.}
During the final optimization phase, the joint configuration $Q^*$ is retrieved by minimizing a composite energy function. The weighting factors ($\lambda$) balancing the individual energy terms are empirically set to ensure stable convergence. The individual energy terms and their respective weights are defined as follows:

\begin{itemize}
    \item \textbf{Contact Distance ($E_{dist}$, $\lambda_d = 1.0$):} To encourage precise alignment between the predicted spatial contact points $\hat{\calC}(W)$ and the gripper's geometry, we minimize the squared Euclidean distance between each target contact point and the closest point on its assigned kinematic link:
    \begin{equation}
        E_{dist} = \sum_{p \in \hat{\calC}(W)} \min_{h \in \calH_{l(p)}(Q)} \| p - h \|_2^2
    \end{equation}
    where $\calH_{l(p)}(Q)$ represents the subset of handprint points belonging to the specific link $l(p)$ assigned to the contact point $p$.

    \item \textbf{Object Penetration ($E_{pen}$, $\lambda_p = 0.5$):} To ensure physically plausible grasps, we explicitly penalize gripper points that penetrate the target object's volume. Utilizing the object's Signed Distance Field (SDF), denoted as $\Phi_{\calO}(\cdot)$, this term is formulated as:
    \begin{equation}
        E_{pen} = \sum_{h \in \calH(Q)} \max(0, - \Phi_{\calO}(h))
    \end{equation}

    \item \textbf{Self-Penetration ($E_{spen}$, $\lambda_s = 0.01$):} We prevent self-collisions between different fingers or parts of the robotic hand by enforcing a minimum spatial safety threshold $\tau$ (set to $0.025$\,m) between disjoint kinematic links $i$ and $j$:
    \begin{equation}
        E_{spen} = \sum_{i,j} \max(0, \tau - \operatorname{dist}(k_i(Q), k_j(Q)))^2
    \end{equation}
    where $k_i(Q)$ denotes the spatial centroid of the bounding box for link $i$ in a given configuration $Q$.

    \item \textbf{Joint Limits ($E_{j}$, $\lambda_j = 1.0$).} We strictly constrain the optimized joint angles to remain within the physical hardware's kinematic limits, defined by $[Q_{min}, Q_{max}]$:
    \begin{equation}
        E_{j} = \| \max(0, Q - Q_{max}) \|_2^2 + \| \max(0, Q_{min} - Q) \|_2^2
    \end{equation}
    
    \item \textbf{Repulsive Energy Term ($E_{rep}$, $\lambda_r = 0.01$):}
    During the energy-based joint optimization, we aim to align the gripper to the predicted contacts while strictly preventing unintended parts of the hand from resting on the object, which would violate the requested contact topology. To achieve this, we introduce the repulsive energy term $E_{rep}$, which penalizes object proximity for all gripper handprint points belonging to non-active regions:
    \begin{equation}
        E_{rep} = \sum_{h \in \calH(Q) \setminus \calH_{l(p)}(Q)} \text{max}(0, \delta - \text{dist}(h, \tilde{\calO}))    
    \end{equation}
    We enforce a minimum safety margin of $\delta = 0.005$ m.
\end{itemize}

\section{Topology-Conditioned Pose Sampling Heuristic}
\label{supp_mat:sec:pose_sampling}

\begin{figure}[!htbp]
    \centering
    \includegraphics[width=0.8\textwidth]{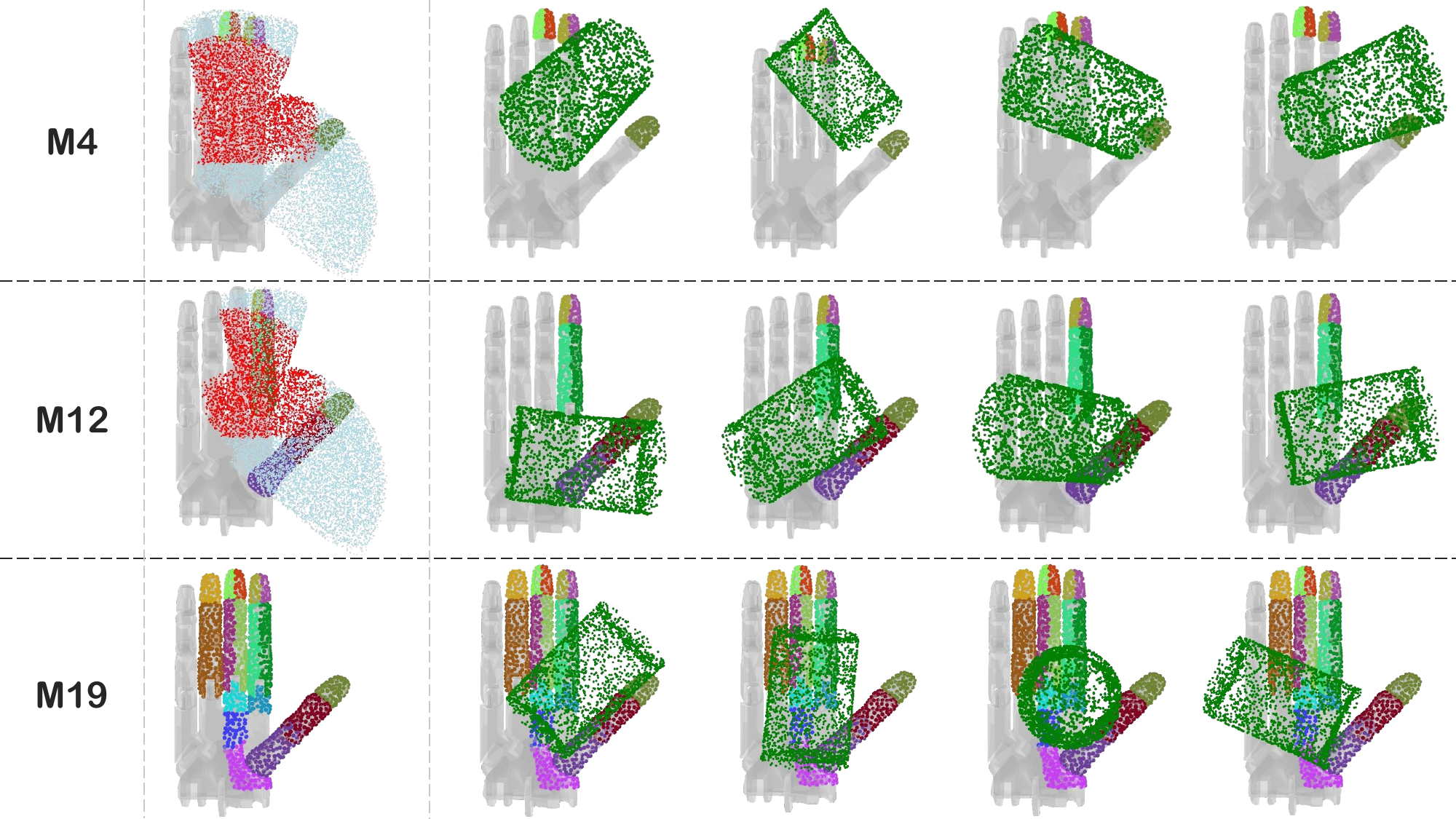}
    \caption{\textbf{Topology-Conditioned Pose Sampling.} For specific grasp topologies (such as M4 and M12), the initial object pose is sampled within a restricted kinematic region. \textbf{Middle:} The template's full active sub-workspace is shown in light blue, while the truncated sub-workspace -- filtered for reachability and palm clearance -- is highlighted in red. \textbf{Right:} Examples of the initialized object point cloud $\tilde{\calO}$ (green) successfully placed within this feasible region after the sampled spatial transformation.}
    \label{supp_mat:fig:rt_sampling}
\end{figure}

As discussed in the main text, \acronyme treats the global 6D grasp pose as an external prior. To autonomously evaluate the framework without relying on external task planners, we sample diverse candidate object poses $(R, \mathbf{t})^{-1}$ using a topology-conditioned heuristic tailored to the specific active zones of the requested grasp template. Figure~\ref{supp_mat:fig:rt_sampling} illustrates the following sampling strategy for three distinct contact topologies.

\noindent\textbf{Palm-Constrained Topologies.}
For power grasps and contact topologies requiring palm contact, we adopt the spatial heuristic from~\cite{merand2026goag, liu2021synthesizing, li2023gendexgrasp}. We uniformly sample approach translations and orientations on the object's convex hull, directed toward its volumetric center. The object is then transformed into the canonical gripper frame via the inverse transformation $(R, \mathbf{t})^{-1}$.

\noindent\textbf{Distal and Precision Topologies.}
For grasps explicitly excluding the palm (\eg fingertips only), standard convex hull sampling often yields kinematically unreachable configurations. Instead, we sample the object pose $(R, \mathbf{t})^{-1}$ directly within a restricted kinematic sub-workspace of the target template's active zones. 
To ensure the object is placed in a feasible region away from the palm, we systematically truncate the sub-workspace using three geometric constraints: (1) a radial distance threshold from a predefined workspace center to exclude out-of-reach boundary points, (2) a directional half-space filter ensuring points lie strictly in front of the palm's normal axis, and (3) a minimum height threshold along the z-axis to avoid the lower hand geometry.
Furthermore, to guarantee topological validity and prevent trivial local minima during the energy-based joint optimization, we explicitly enforce a strict 1 cm clearance margin between the object geometry and the palm.

\section{Automated Grasp Classification Metric}
\label{supp_mat:sec:type_classification}

This section details the mathematical formulation of our contact topology recognition metric.

\noindent\textbf{Contact Extraction and Zone Mapping.}
First, we extract the subset of gripper points in contact with the object, $\calC_{\text{grip}}(\calH(Q))$, using the aligned distance formulation with an empirically chosen proximity threshold of $\epsilon = 0.8$. To compare these physical contacts against the taxonomy $\calT$, we map the raw contact points back to their semantic zone identifiers.
Let $I_{\text{obs}} = \{ i \in \calI \mid \mathbf{h}_i(Q) \in \calC_{\text{grip}}(\calH(Q)) \}$ be the indices of the gripper points currently touching the object. Using the surjective mapping $\zeta$, we define the observed active zones ($A_{\text{obs}}$) as the set of unique physical regions engaged in the grasp, and the ground-truth required zones ($A_m$) extracted from the target template $\calA_m$ (explicitly excluding the label $0$):
\begin{equation}
    A_{\text{obs}} = \{ \zeta(i) \mid i \in I_{\text{obs}} \} \qquad \text{and} \qquad A_m = \{ \calA_m(i) \mid i \in \calI \} \setminus \{0\}
\end{equation}

\noindent\textbf{Asymmetric Tversky Index.}
We evaluate the match between the generated grasp and a specific contact topology $m$ by computing a similarity score $s_m$ based on the Tversky index~\cite{tversky1977features}:
\begin{equation}
    s_m = \frac{|A_{\text{obs}} \cap A_m|}{|A_{\text{obs}} \cap A_m| + w_1 |A_{\text{obs}} \setminus A_m| + w_2 |A_m \setminus A_{\text{obs}}|}
\end{equation}
where $|\cdot|$ denotes set cardinality. We set the weighting parameters to $w_1 = 2.0$ and $w_2 = 0.5$.
This asymmetric penalization, determined empirically, is critical: it punishes extra contact regions ($|A_{\text{obs}} \setminus A_m|$), which typically denote clumsy or unintended enveloping power grasps, much more strictly than missing contacts ($|A_m \setminus A_{\text{obs}}|$). \\
Any grasp failing to reach a similarity threshold of $s_{m} \ge 0.5$ is strictly classified as 'unknown'. Given our asymmetric Tversky weights, $w_1$ and $w_2$, $0.5$ represents the mathematical tipping point where the number of correctly aligned contact zones strictly outweighs the heavily penalized hallucinated contacts (e.g., a failed precision pinch collapsing into a power grasp), ensuring robust rejection of degenerated topologies.

\section{Standard Evaluation Metrics Protocol}
\label{supp_mat:sec:metrics}

As mentioned in the main text, \acronyme utilizes the standard evaluation protocols widely established in~\cite{merand2026goag, wei2024d, li2023gendexgrasp, liu2021synthesizing} to measure baseline physical performance.

\noindent\textbf{Success Rate (SR).}
We evaluate the physical stability of the synthesized grasps using the Isaac Gym simulator~\citesupp{makoviychuk2021isaac}. The target object (standardized to a mass of 100 g) and the gripper are initialized in the optimized configuration $Q^*$. We sequentially apply external perturbations on the object along the $\pm x, y, z$ axes for one second each. A grasp is considered successful if the object's translation deviates by less than 2 cm from its initial pose. Furthermore, any grasp exhibiting severe interpenetration (contact forces exceeding 500 N) is strictly classified as a failure to penalize unrealistic physical states. \\
It is crucial to note that the theoretical force-closure validation performed during inference serves only as a rapid geometric estimation based on idealized point contacts. It offers no absolute guarantees regarding final physical stability, as the Isaac Gym simulation rigorously tests the complex reconciliation of these points against the complete object geometry, strict kinematics, and the torque saturation limits of the gripper's simulated actuators when compensating for external perturbation forces.

\noindent\textbf{Generation Speed.}
We report the average computational time (in seconds) required to generate a single grasp, calculated over 100 generation attempts. This metric strictly isolates the network inference and the energy-based joint optimization time, explicitly excluding the heavy physics simulation overhead of Isaac Gym.

\noindent\textbf{Spatial Diversity.}
We quantify generative spatial diversity by calculating the standard deviation of the gripper translation ($\mathbf{t}$), orientation ($R$), and joint values ($Q$). Because the initial global poses $(R, \mathbf{t})$ are derived from our topology-conditioned sampling heuristic, the diversity in $\mathbf{t}$ and $R$ directly reflects the method's spatial reachability and adaptability across various object geometries.

\section{Detailed Experimental Protocols and Baselines}
\label{supp_mat:sec:baselines}

To ensure a rigorous and fair evaluation, this section provides the extended protocols and baseline configurations.

\noindent\textbf{Taxonomy-Unaware Evaluation Protocol.}
To evaluate the inherent functional bias of unconditioned grasp planners, we compare \acronyme against four state-of-the-art baselines: DFC~\cite{liu2021synthesizing}, GenDexGrasp~\cite{li2023gendexgrasp}, DRO-Grasp~\cite{wei2024d}, and GOAG~\cite{merand2026goag}.
For this experiment, we utilize the test subset from the MultiDex dataset, comprising 10 objects from ContactDB~\citesupp{brahmbhatt2019contactdb} and YCB~\cite{calli2015ycb}. Each baseline was tasked with generating exactly 100 grasps per object. To ensure a fair comparison, the baselines are deployed as follows:
\begin{itemize}
    \item \textbf{DFC:} as DFC does not require training, we utilize pre-generated grasps from the CMapDataset. These poses were originally synthesized via DFC and subsequently refined by~\cite{li2023gendexgrasp} through a post-filtering process to ensure geometric validity.
    \item \textbf{GenDexGrasp:} We use the official pre-trained checkpoint (trained on multi-hand data) with default hyperparameters. We first infer the contact maps for the test objects, which are then passed into the authors' optimization framework to derive the final grasp poses.
    \item \textbf{DRO-Grasp:} We select the most performant variant, which includes config\-uration-invariant pre-training. We retain all default hyperparameters but explicitly deactivate their simple grasp controller to ensure a fair, purely kinematic comparison.
    \item \textbf{GOAG:} We train GOAG from scratch on the Shadow Hand kinematics using the default hyperparameters, and generate grasps following the standard inference pipeline.
\end{itemize}
All successfully generated, physically stable grasps were subsequently fed into our automated classification pipeline (Sec.~\ref{subsec:exp_setup}, Supp. Mat.~\ref{supp_mat:sec:type_classification}) to evaluate their Topology Compliance (\textbf{TC}), Entropy and Spatial Diversity. 

Because these baselines do not take a functional condition as input, this test strictly isolates their natural generative distribution, exposing the severe mode collapse toward power grasps driven by standard physics-based optimization.
Furthermore, while \acronyme achieves a higher \textbf{TC} than the baselines, the absolute scores remain relatively low across all methods. This shared ceiling suggests that \textbf{TC} is heavily bottlenecked by the specific object set utilized; the limited geometric diversity of these test objects naturally restricts the subset of physically viable grasps, making certain functional topologies kinematically unattainable regardless of the generative method.

\noindent\textbf{Taxonomy-Aware Evaluation Protocol.}
To evaluate the precise functional control of \acronyme, we conducted an extensive comparison against Dexonomy~\cite{chen2025dexonomy} on the full test set of the DexGraspNet~\cite{wang2023dexgraspnet} dataset, which comprises 1,126 geometrically diverse objects. We selected Dexonomy~\cite{chen2025dexonomy} as our sole baseline for this experiment, as it currently stands as the first and only state-of-the-art framework capable of taxonomy-conditioned generative grasp synthesis. While other recent methods tackle dexterous grasping, they rely on assumptions orthogonal to our object-agnostic setting. For instance, functional grasp transfer and retargeting methods, such as FunGrasp~\cite{huang2025fungrasp} and others~\citesupp{yang2022oakink, antotsiou2018task}, operate in a fundamentally different problem space; they inherently require explicit human grasp demonstrations or a dense spatial prior—such as a source human hand mesh—to initialize their optimization. Furthermore, unconditioned frameworks like AnyDexGrasp~\cite{fang2025anydexgrasp} lack semantic topology control, while methods like OmniDexVLG~\cite{zhang2025omnidexvlg} rely on 2D VLM supervision and are currently publicly unavailable. In contrast, \acronyme and Dexonomy~\cite{chen2025dexonomy} function as true grasp planners, synthesizing functionally compliant grasps from scratch relying purely on raw object geometry and a discrete semantic label. \\
\underline{Taxonomy Mapping and Grouping:} Dexonomy~\cite{chen2025dexonomy} natively outputs grasps categorized under the classic Feix taxonomy~\cite{feix2015grasp}. To ensure a direct one-to-one semantic comparison, we mapped their analytically computed dataset to our contact-based topologies (illustrated in Figure~\ref{fig:grasp_taxonomy_contact} of the main paper). Crucially, this evaluation space ensures an unbiased comparison. By focusing exclusively on contact regions, the Feix taxonomy naturally reduces to our adopted Gonzalez framework (\eg, F12/F13 $\rightarrow$ M6). While Dexonomy couples joint configurations with contact semantics, \acronyme relies strictly on contact topologies. Because our metrics ($\mathbf{TC}$, $\mathbf{H}_{TC}$) measure semantic compliance purely through these spatial contact assignments, the two taxonomies are structurally equivalent for this assessment. Finally, we explicitly grouped the evaluated grasps into three functional categories to analyze performance across different dexterity levels: \textit{Power} grasps (enveloping, high stability), \textit{Precision} grasps (fingertip manipulation, low stability), and \textit{Object-Specific} grasps (tool use). \\
\underline{Handling Geometric Incompatibility:} A critical challenge in large-scale grasp evaluation is that not all contact topologies are geometrically possible on all objects (\eg it is physically impossible to execute a tiny fingertip precision pinch on a massive, smooth sphere). If an object cannot support a specific contact topology, penalizing the network for failing to generate it skews the metric and misrepresents the model's actual generative capability. \\
To ensure a strictly fair comparison, we implemented a geometric compatibility filter. Any object-topology pair that yielded a $0\%$ success rate across all generation attempts was deemed fundamentally geometrically incompatible and explicitly excluded from the final average calculations. \\
Following this rigorous filtering process, \acronyme successfully covers an average of $80.14\%$ of the objects per requested contact topology, which closely and fairly matches Dexonomy's coverage of $81.05\%$. Consequently, the Success Rate and Topology Compliance metrics reported in Table~\ref{tab:tax_aware_results} strictly reflect \acronyme's superior generative control on valid geometries, rather than an artifact of object selection.

\section{Topology Compliance Analysis and Simulation Bottlenecks}
\label{supp_mat:sec:tc_analysis}

To further understand the discrepancy between requested and effective contact topologies observed in the main paper, we analyze \acronyme's \textbf{TC} at different stages of the generation pipeline (Table~\ref{supp_mat:tab:results_tc}).

\begin{table}[h]
    \centering
    \resizebox{0.5\textwidth}{!}{
    \begin{tabular}{c|c|c|c|c}
        \toprule
        \textbf{Label}
        & \textbf{Eval.}
        & \multirow{2}{*}{$\textbf{H}_{SR}$} & \multirow{2}{*}{\textbf{TC (\%)}} & \multirow{2}{*}{$\textbf{H}_{TC}$}
        \\ \textbf{Consistency} & \textbf{Isaac} & & & 
        \\  \midrule
        \cmark & \cmark & \multirow{2}{*}{0.96} & 17.18 & 0.84
        \\
        \cmark & \xmark &  & 21.92 & 0.53
        \\ \hline
        \xmark & \cmark & \multirow{2}{*}{0.94} & 14.45 & 0.81
        \\
        \xmark & \xmark &  & 19.34 & 0.50
        \\
        \bottomrule
    \end{tabular}
    }
    \caption{Topology compliance ablation.}
    \label{supp_mat:tab:results_tc}
\end{table}

\noindent\textbf{Value of Topological Filtering.}
Comparing the top and bottom halves of the table demonstrates the value of topological filtering. By rejecting geometrically incompatible templates before the energy-based optimization, the pipeline prevents the optimizer from coercing the hand into unnatural configurations that would ultimately fail in simulation, thereby improving the overall semantic quality of the successful grasps.

\begin{figure}[!htbp]
    \centering
    \includegraphics[width=0.5\textwidth]{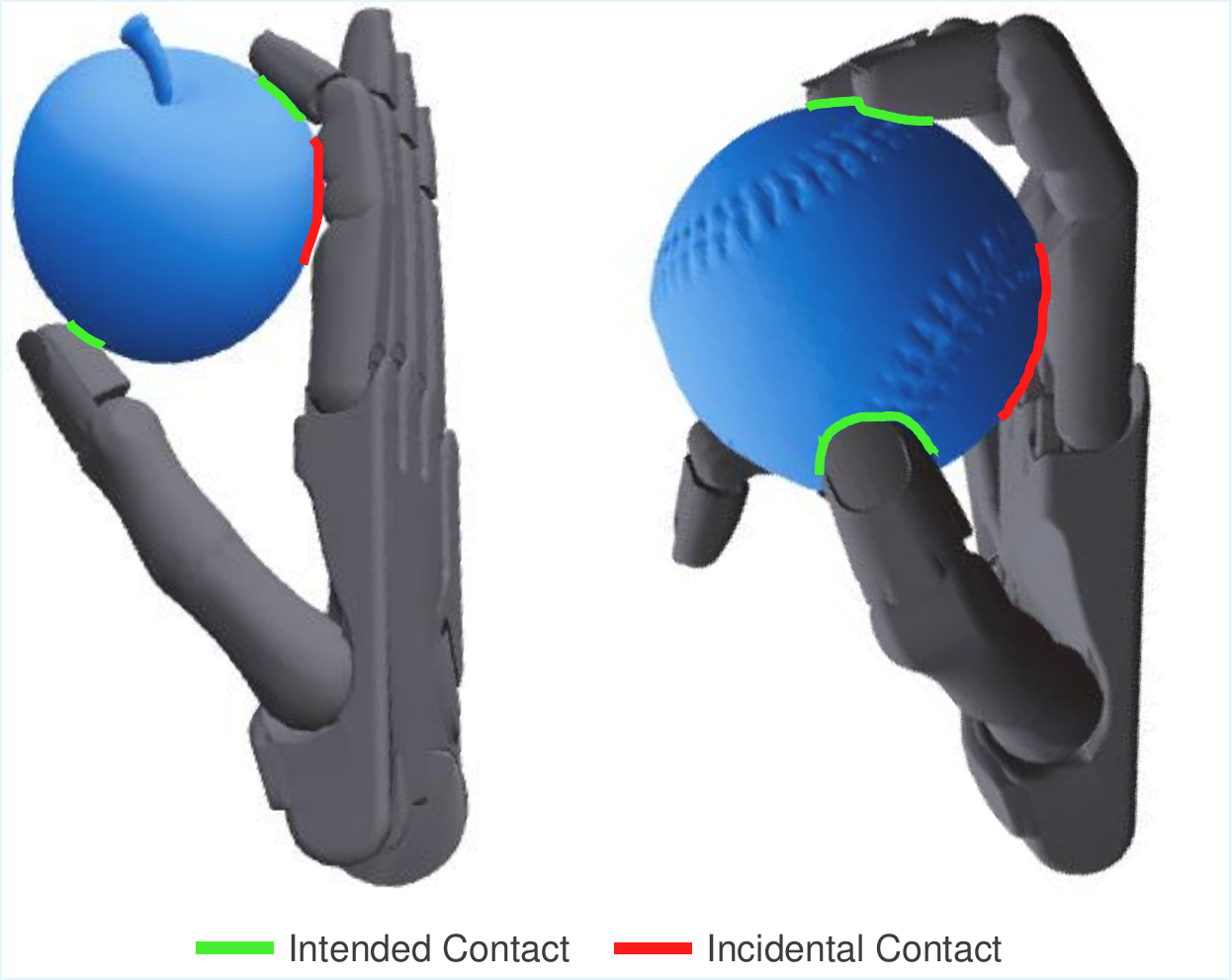}
    \caption{\textbf{Illustration of metric-induced misclassification.} While \acronyme strictly respects the target contact topology, the kinematic optimization may result in \textbf{\red{incidental contacts}} where an adjacent phalanx rests against the object surface. Despite the repulsive term in our optimization energy function, these phalanges often cannot be pushed away due to inherent kinematic constraints. For instance, the left grasp illustrates an intended M3 pinch reclassified as M12, while the right shows an M4 grasp reclassified as M15. Although the \textbf{\green{intended contacts}} are successfully achieved and the grasps remain physically stable, these incidental contacts trigger a strict reclassification by our automated pipeline.}
    \label{supp_mat:fig:incidental_contact}
\end{figure}

\noindent\textbf{The Physics-Based Metric Drop.}
The most striking shift occurs between the pre-physics (\xmark\xspace Eval. Isaac) and post-physics evaluations (\cmark\xspace Eval. Isaac). Prior to the Isaac Gym simulation, the pipeline achieves its highest absolute \textbf{TC} but exhibits a remarkably low semantic entropy ($\textbf{H}_{TC}$). This indicates that the purely geometric, energy-based optimizer frequently converges on configurations that technically satisfy the target contact masks but lack true physical stability. Once subjected to gravity and external perturbations, a significant portion of these superficial grasps naturally fails. While this physics-based filtering prunes weak configurations and drastically rebalances the distribution (restoring $\textbf{H}_{TC}$), it also causes a drop in the raw \textbf{TC}.
This drop exposes a sensitivity to the rigid definition of our similarity score formula. During simulation, fingers naturally shift slightly along the object's local curvature to achieve a stable force-closure. Because our metric enforces strict mathematical boundaries, these minor, functionally viable adjustments often lead to misclassifications. Specifically, despite the presence of a repulsive term in the optimization energy function, certain phalanges cannot be pushed away from the object surface due to inherent kinematic constraints, such as fixed finger lengths or joint limits. In these scenarios, the optimization weighting factors privilege the primary contact required by the topology over the repulsive term, leading to incidental contacts. This phenomenon frequently results in intended precision pinches (M2 or M3) being penalized and reclassified as M12 due to an adjacent phalanx resting too closely to the surface -- as illustrated in Figure~\ref{supp_mat:fig:incidental_contact}. 
Consequently, while the final generated grasps are physically stable, this rigid metric provides an incomplete picture of functional success, as it fails to capture whether the intended core contacts were actually achieved.

\begin{figure}[!htbp]
    \begin{subfigure}{\textwidth}
        \centering
        \includegraphics[width=\textwidth]{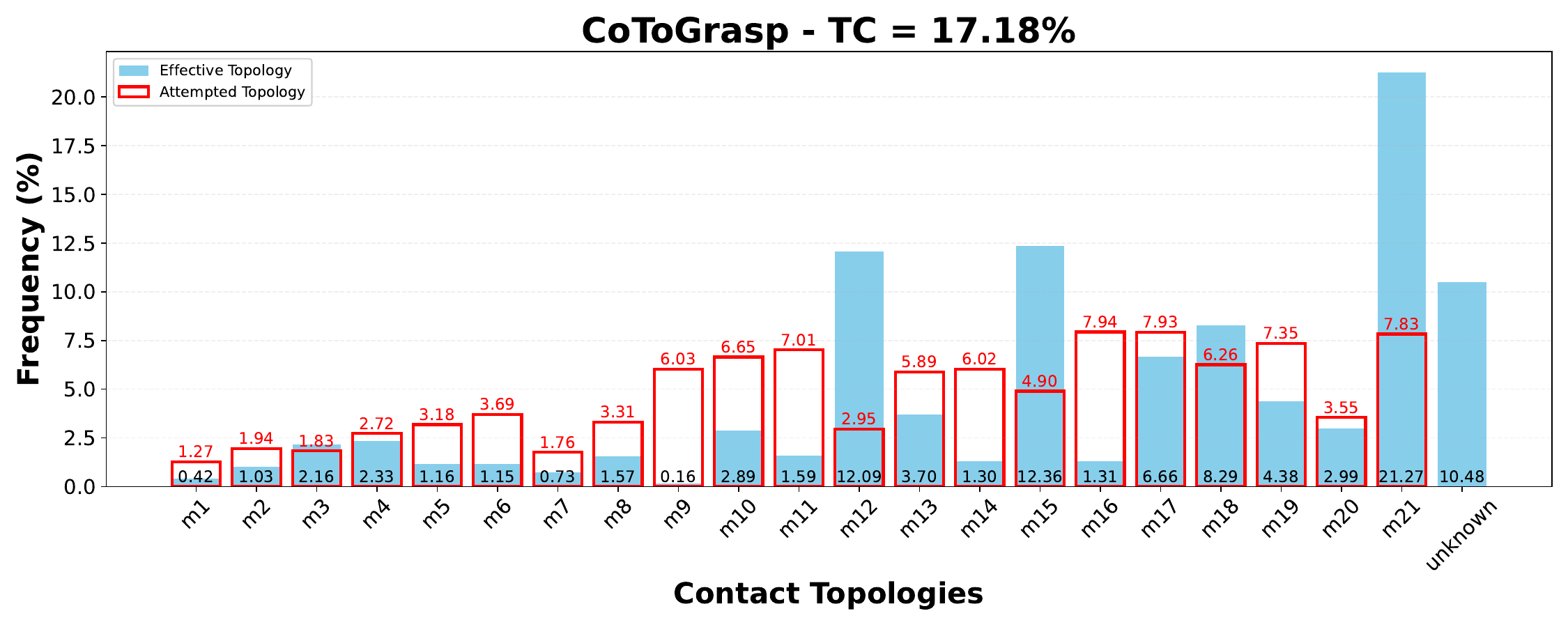}
    \end{subfigure}
    \begin{subfigure}{\textwidth}
        \centering
        \includegraphics[width=\textwidth]{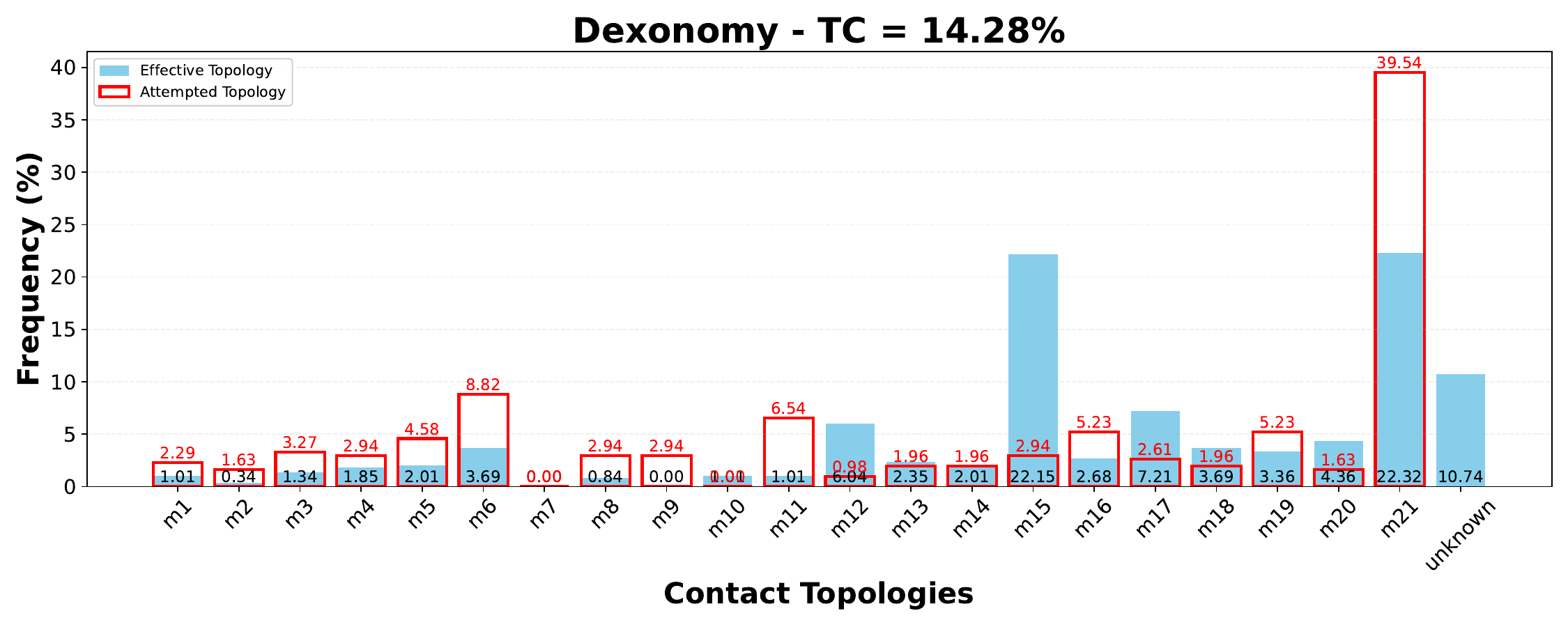}
    \end{subfigure}
    \caption{Histogram comparing the frequencies of \textbf{\lightblue{effective topologies}} and \textbf{\red{attempted topologies}} (as defined Sec.~\ref{subsec:metrics}) among all stable grasps generated by \acronyme (top) and Dexonomy~\cite{chen2025dexonomy} (bottom).}
    \label{supp_mat:fig:type_freq}
\end{figure}

\noindent\textbf{Global Topological Distribution.}
Beyond the specific pipeline bottlenecks, Figure~\ref{supp_mat:fig:type_freq} provides a comprehensive visual breakdown of the attempted versus effective topologies across all 21 topologies. As illustrated, while both methods experience the aforementioned metric-induced shifts during simulation, \acronyme maintains a substantially more balanced and faithful distribution of functional grasps (\textbf{TC} $= 17.18\%$) compared to the Dexonomy baseline (\textbf{TC} $= 14.28\%$).

\noindent\textbf{Object-Level Geometric Complexity.}
To quantify geometric difficulty and evaluate the inherent trade-off between strict semantic control and geometric flexibility, we analyzed performance grouped by object convexity ($c = V_{obj}/V_{hull}$). As detailed in Table~\ref{supp_mat:tab:object_level_analysis}, we define \textit{SR Retained} as the comparison of Success Rates (SR) between non-convex ($c < 0.4$) and convex ($c > 0.9$) objects. When transitioning to these challenging geometries, \acronyme demonstrates remarkable robustness, retaining 58.72\% of its physical SR. In contrast, Dexonomy's rigid templates retain only 37.42\% of their performance, and the unconditioned GOAG drops to 28.09\%. This performance gap widens on objects with severe concavities ($c < 0.4$, representing roughly 4\% of the dataset). On these hardest geometries, \acronyme achieves an 18.30\% SR, outperforming Dexonomy's 12.05\%. Strikingly, \acronyme's semantic compliance (\textbf{TC}) actually increases to 19.17\% on these objects, whereas Dexonomy's \textbf{TC} collapses to 8.94\%. These results prove that our contact-centric approach thrives on leveraging complex local features to anchor semantics. While strict semantic control typically limits geometric flexibility, \acronyme pushes this boundary significantly further than planners relying on rigid, joint-coupled templates, which systematically fail or abandon the semantic query entirely on complex shapes.

\begin{table}[htbp]
    \centering
    \small
    \setlength{\tabcolsep}{4pt}
    \begin{tabular}{l|c|cc}
        \toprule
        \textbf{Object Complexity} & \textbf{Metric} & \textbf{\acronyme} & \textbf{Dexonomy~\cite{chen2025dexonomy}} \\ \hline
        Convex $\rightarrow$ Non-Convex & SR Retained & 58.72\% & 37.42\% \\ \hline
         \multirow{2}{*}{\begin{tabular}[c]{@{}c@{}} Severe Concavities \\ ($c < 0.4$, $\sim 4\%$ data) \end{tabular}} & SR & 18.30\% & 12.05\% \\
         \cline{2-4}
         & TC & 19.17\% & 8.94\% \\
         \bottomrule
    \end{tabular}
    \caption{\textbf{Object-Level Analysis.} Evaluating performance across geometric complexity ($c = V_{obj}/V_{hull}$). \acronyme shows superior retention of both physical stability (SR) and semantic compliance (TC) on challenging non-convex objects.}
    \label{supp_mat:tab:object_level_analysis}
\end{table}

\section{Real-World Experiments}
\label{supp_mat:sec:real_world}

\noindent\textbf{Execution Protocol:}
For a given synthesized grasp $(R, \mathbf{t}, Q)$, we implemented a strict execution pipeline.
To ensure collision-free trajectories and safe hardware operation, all motions were initially planned and validated within a ROS2 digital twin using the MoveIt~\citesupp{chitta2012moveit} motion planning framework.
The execution sequence begins by computing an approach pose, defined by translating the target 6D pose $(R, \mathbf{t})$ backward by 10~cm along the normal vector of the gripper's palm. The robot first navigates to this approach pose with the fingers fully extended. Subsequently, the arm linearly interpolates to the target spatial pose $(R, \mathbf{t})$, and the fingers are actuated to converge on a squeezed configuration $Q_{s}^{*}$ based on the optimized joint configuration $Q^*$. $Q_{s}^{*}$ is computed such as finger links involved in the grasps are closer to the object's center of mass, similarly as~\cite{wei2024d, chen2025dexonomy}.
To empirically verify the physical stability of the grasp against gravity, the manipulator lifts the object 15~cm directly above the tabletop, holds it statically for 3 seconds, and finally opens the hand to release the object.

\noindent\textbf{Discussion and Limitations.}
The execution of synthesized precision grasps on physical hardware exposes the fundamental mismatch between deterministic kinematic planning and the stochastic physics of mechanical contact. While standard position-control frameworks (\eg OMPL pipelines in MoveIt) are effective for coarse manipulation, they fail to maintain the integrity of complex contact topologies. We observed that relying on a simple linearly interpolated squeezed configuration $Q_{s}^{*}$ is critically insufficient; because $Q_{s}^{*}$ is derived from geometric distances to the object's barycenter, digits move at uniform velocities regardless of external reaction forces. This inevitably results in asynchronous contact, where leading fingers strike the surface milliseconds early, generating unbalanced moments that displace the object before force-closure is achieved. Consequently, physical contact points drift from predicted optimal zones, misaligning force vectors with the intended functional semantics. Furthermore, current evaluation benchmarks -- often restricted to tabletop environments -- contradict the objective of omnidirectional grasp synthesis. Such environments artificially inflate the success of grasps that may only be viable in a bimanual setup or when the object is presented in a specific, pre-constrained 6D pose. To maintain fidelity to simulated topologies, future work must transition toward contact-aware interaction paradigms, such as hybrid force/position control or tactile-reactive policies~\citesupp{jahanshahi2024review}.

\section{Qualitative Results: \acronyme Visualization}
\label{supp_mat:sec:grasp_visu}

\begin{figure}[!htbp]
    \centering
    \includegraphics[width=\textwidth]{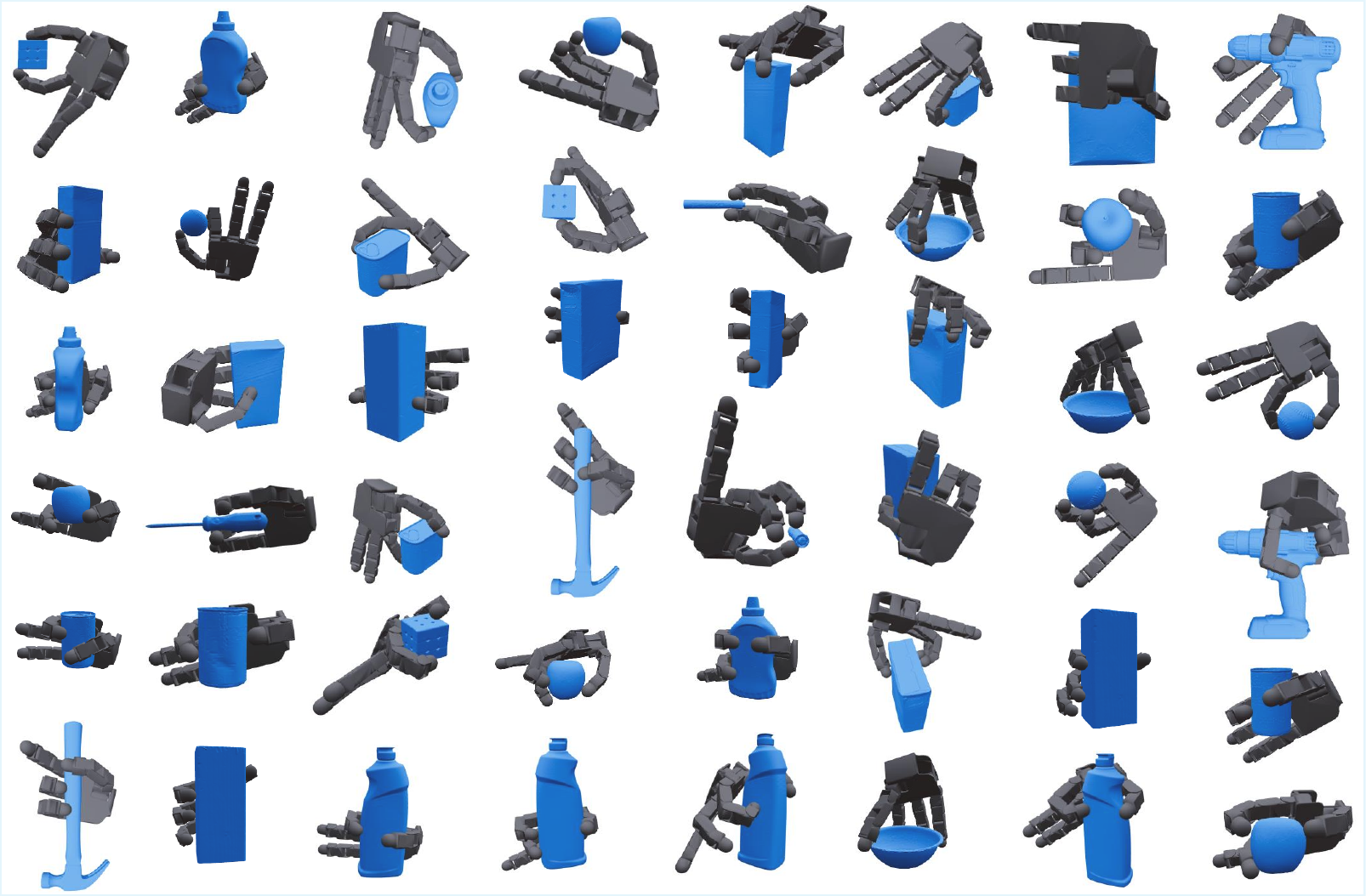}
    \caption{\textbf{Qualitative Synthesis Gallery on the Allegro Hand.} Synthesized grasp configurations for a diverse subset of the YCB object dataset.}
    \label{supp_mat:fig:grasps_allegro}
\end{figure}

\begin{figure}[!htbp]
    \centering
    \includegraphics[width=\textwidth]{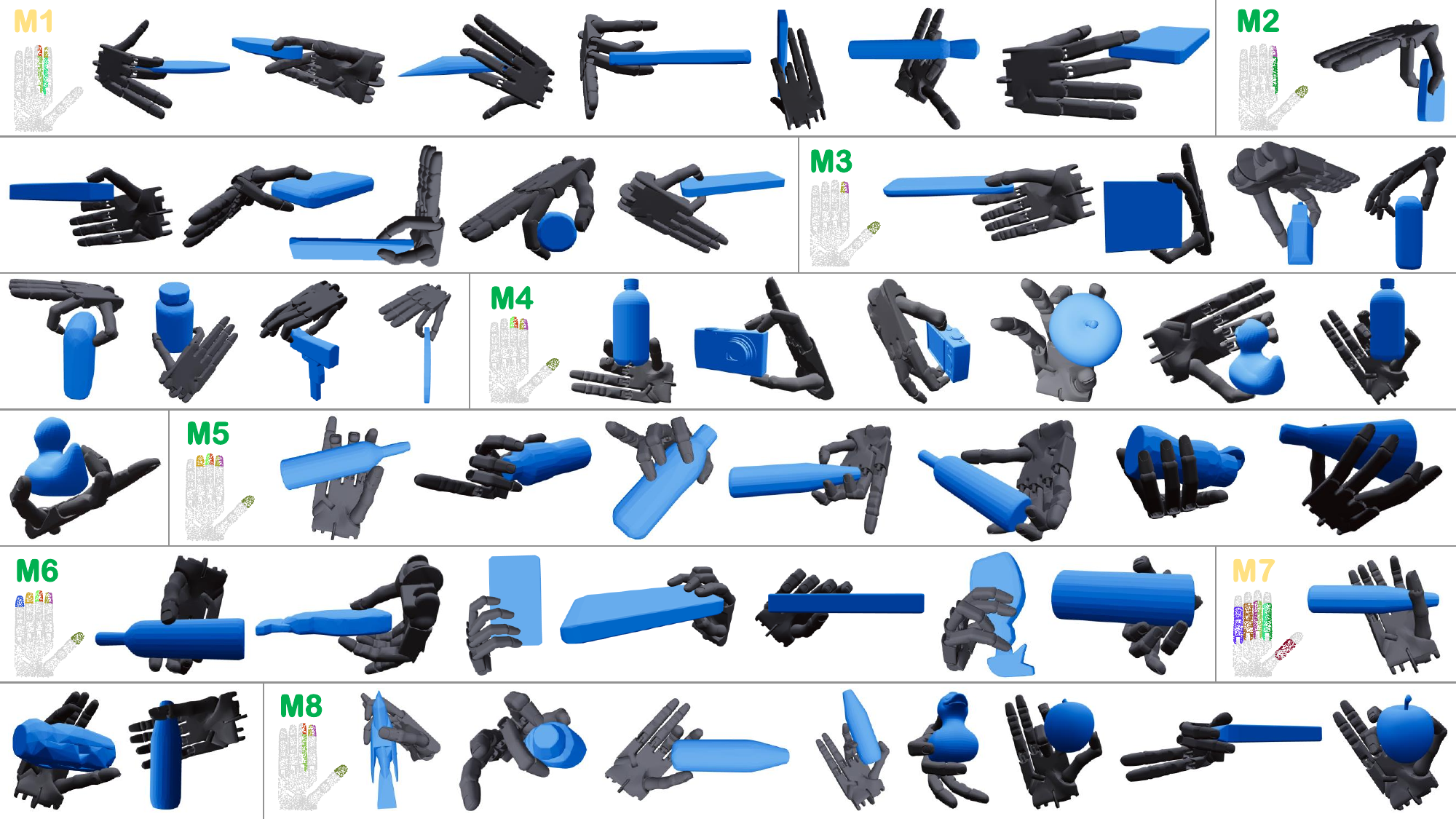}
    \includegraphics[width=\textwidth]{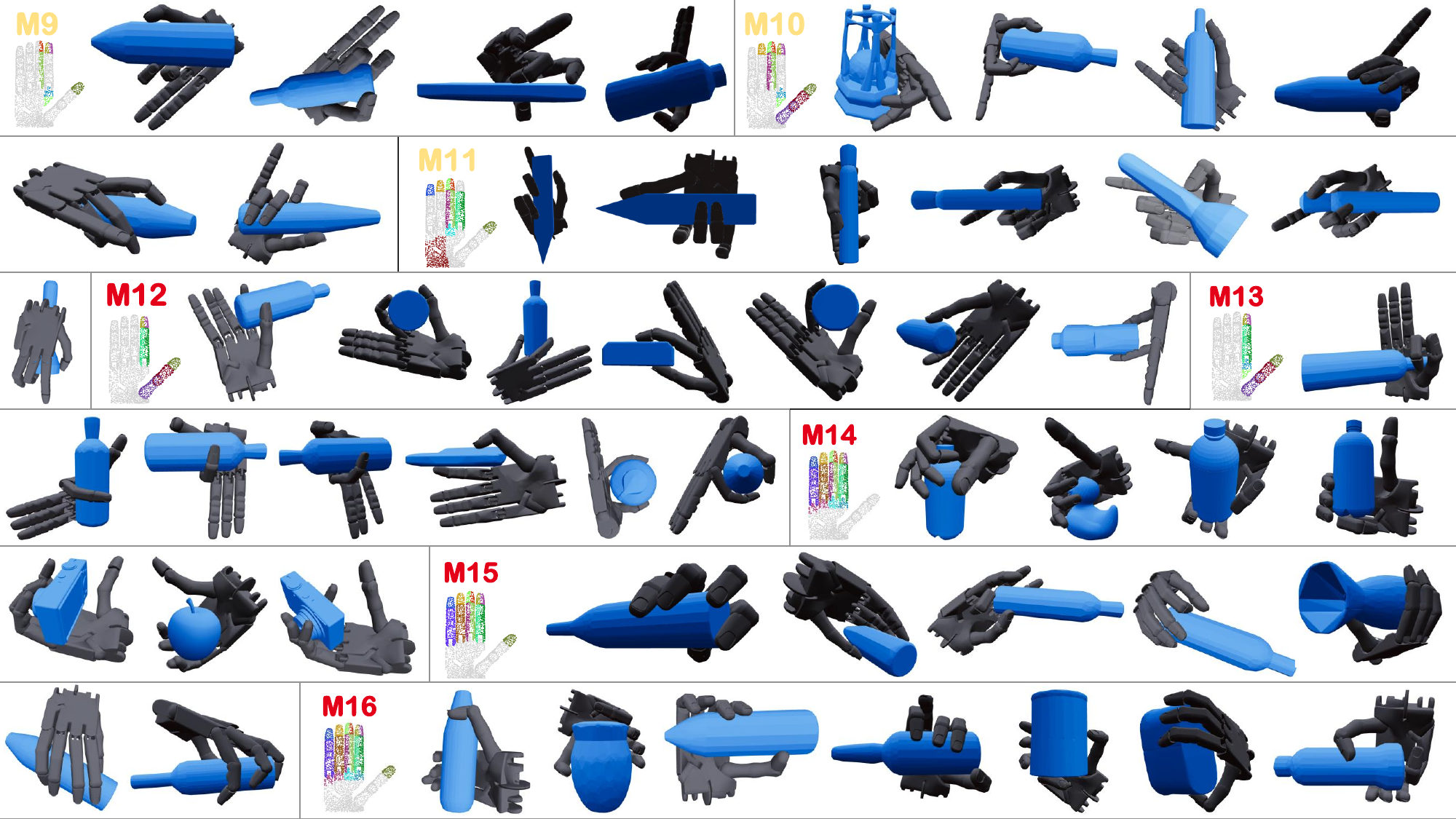}
    \includegraphics[width=\textwidth]{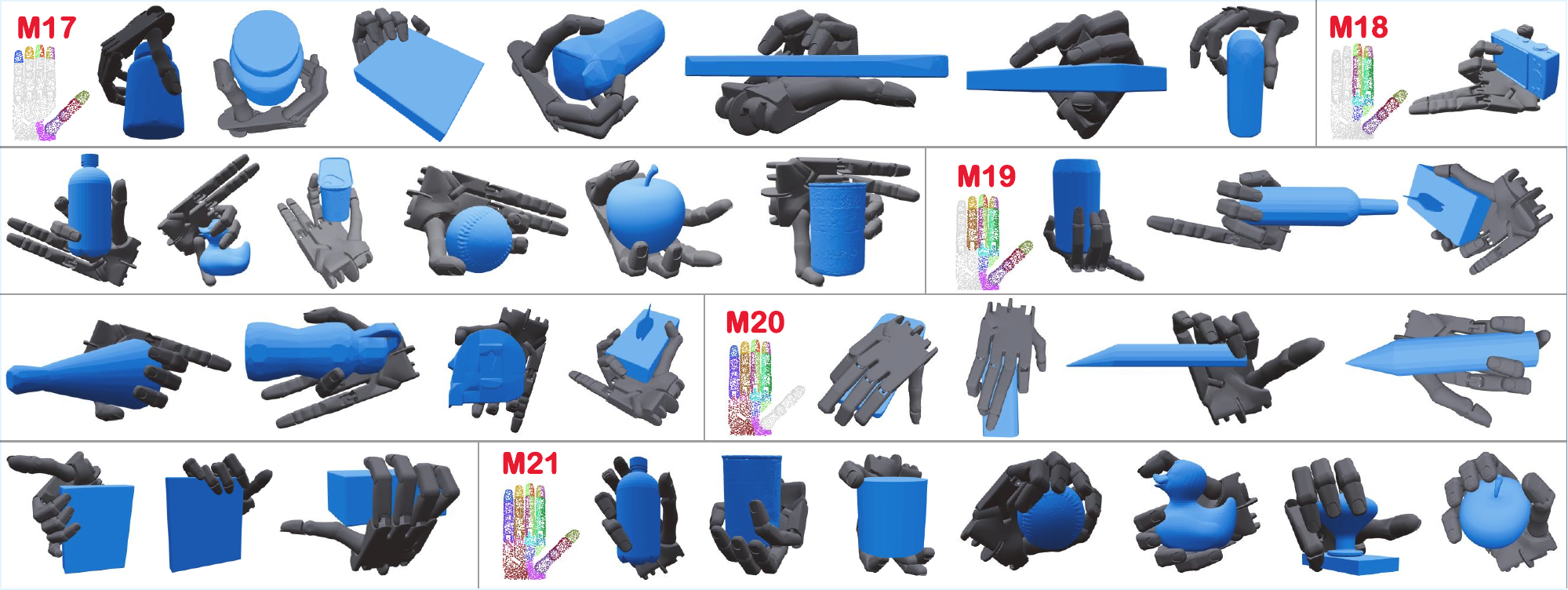}
    \caption{\textbf{Topological Clustering of Shadow Hand Grasps.} Grasps grouped by contact topology (M1-M21), demonstrating consistent semantic alignment across varied object classes. }
    \label{supp_mat:fig:grasps_shadow}
\end{figure}

Figure~\ref{supp_mat:fig:grasps_allegro} and Figure~\ref{supp_mat:fig:grasps_shadow} present qualitative results across a diverse set of YCB and DexGraspNet objects, highlighting the framework's ability to strictly adhere to the intended contact topology. Rather than merely reaching for the object's center of mass, the synthesized configurations demonstrate precise alignment between the fingers and specific semantic contact zones. As shown in Figure~\ref{supp_mat:fig:grasps_shadow}, when grasps are grouped by their topological identifiers (M1-M21), the planner consistently recovers the prescribed contact manifolds. This adherence ensures that the resulting configurations maintain the intended grasp type -- whether a delicate fingertip pinch or a complex multi-digit wrap -- effectively preserving the functional semantics of the interaction across both convex and non-convex surfaces.

\bibliographystylesupp{splncs04}
\bibliographysupp{main}

\end{document}